\documentclass{article}

\usepackage{natbib}
\usepackage[margin=3cm]{geometry}
\usepackage[utf8]{inputenc} % allow utf-8 input
\usepackage[T1]{fontenc}    % use 8-bit T1 fonts
\usepackage{url}            % simple URL typesetting
\usepackage{amsfonts}       % blackboard math symbols
\usepackage{nicefrac}       % compact symbols for 1/2, etc.
\usepackage{microtype}      % microtypography
\usepackage{wrapfig}      % microtypography
\usepackage{graphicx}
\usepackage{booktabs}       % professional-quality tables
\usepackage{amsmath}        % math symbols
\usepackage{mathtools}

\usepackage{authblk}    % For prepring authors affiliations

\usepackage{subfig}      %for multiple subfloats

\usepackage{todonotes}

\usepackage{tikz}%
\usetikzlibrary{positioning, calc, decorations.pathreplacing, patterns, backgrounds, shapes.arrows}%
\usepackage{pgfplots}
\DeclareUnicodeCharacter{2212}{−}
\usepgfplotslibrary{groupplots,dateplot}
\pgfplotsset{compat=newest}
\usepackage{environ} % For defining a helper environment in rescaling tikzpicture

\usepackage{bbm}
\usepackage{manyfoot}%

\usepackage{wrapfig}

\usepackage{stmaryrd} % For double brackets
\usepackage{xparse} % For defining optional arguments

\usepackage{listings}%

\usepackage{color}
\definecolor{darkred}{rgb}{0.6,0.0,0.0}
\definecolor{darkgreen}{rgb}{0,0.50,0}
\definecolor{lightblue}{rgb}{0.0,0.42,0.91}
\definecolor{orange}{rgb}{0.99,0.48,0.13}
\definecolor{grass}{rgb}{0.18,0.80,0.18}
\definecolor{pink}{rgb}{0.97,0.15,0.45}

\lstdefinestyle{colored}{ %
  basicstyle=\ttfamily,
  backgroundcolor=\color{white},
  commentstyle=\color{green}\itshape,
  keywordstyle=\color{blue}\bfseries\itshape,
  stringstyle=\color{red},
}
\lstdefinelanguage{PythonPlus}[]{Python}{
  morekeywords=[1]{,as,assert,nonlocal,with,yield,self,True,False,None,} % Python builtin
  morekeywords=[2]{,__init__,__add__,__mul__,__div__,__sub__,__call__,__getitem__,__setitem__,__eq__,__ne__,__nonzero__,__rmul__,__radd__,__repr__,__str__,__get__,__truediv__,__pow__,__name__,__future__,__all__,}, % magic methods
  morekeywords=[3]{,object,type,isinstance,copy,deepcopy,zip,enumerate,reversed,list,set,len,dict,tuple,range,xrange,append,execfile,real,imag,reduce,str,repr,}, % common functions
  morekeywords=[4]{,Exception,NameError,IndexError,SyntaxError,TypeError,ValueError,OverflowError,ZeroDivisionError,}, % errors
  morekeywords=[5]{,ode,fsolve,sqrt,exp,sin,cos,arctan,arctan2,arccos,pi, array,norm,solve,dot,arange,isscalar,max,sum,flatten,shape,reshape,find,any,all,abs,plot,linspace,legend,quad,polyval,polyfit,hstack,concatenate,vstack,column_stack,empty,zeros,ones,rand,vander,grid,pcolor,eig,eigs,eigvals,svd,qr,tan,det,logspace,roll,min,mean,cumsum,cumprod,diff,vectorize,lstsq,cla,eye,xlabel,ylabel,squeeze,}, % numpy / math
}

\lstdefinestyle{colorEX}{
  basicstyle=\small\ttfamily,
  backgroundcolor=\color{white},
  commentstyle=\color{darkgreen}\slshape,
  keywordstyle=\color{blue}\bfseries\itshape,
  keywordstyle=[2]\color{blue}\bfseries,
  keywordstyle=[3]\color{grass},
  keywordstyle=[4]\color{red},
  keywordstyle=[5]\color{orange},
  stringstyle=\color{darkred},
  emphstyle=\color{pink}\underbar,
}
\makeatletter
\newsavebox{\measure@tikzpicture}
\NewEnviron{scaletikzpicturetowidth}[1]{%
  \def\tikz@width{#1}%
  \begin{lrbox}{\measure@tikzpicture}%
  \BODY
  \end{lrbox}%
  \pgfmathparse{#1/\wd\measure@tikzpicture}%
  \BODY
}
\makeatother

\newcommand{\ie}{\emph{i.e.\ }}
\newcommand{\eg}{\emph{e.g.\ }}

\newtheorem{remark}{Remark}
\newtheorem{theorem}{Theorem}

\setcitestyle{round,authoryear}
\renewcommand{\vec}{\pmb}
\newcommand{\x}{\vec{x}}
\newcommand{\z}{\vec{z}}

\NewDocumentCommand{\naturalto}{m}{\llbracket #1 \rrbracket}
\newcommand{\real}{\mathbb{R}}
\newcommand{\rkhs}{\mathcal{H}}

\newcommand{\I}{\mathcal{I}}
\newcommand{\normal}{\mathcal{N}}
\newcommand{\dataspace}{\mathcal{X}}
\newcommand{\cluster}{\mathcal{C}}
\newcommand{\dataset}{\mathcal{D}}
\newcommand{\objective}{\mathcal{L}}

\newcommand{\pdata}{p_\text{data}}
\newcommand{\p}{p_\theta}

\NewDocumentCommand{\distance}{o o m}{
  \ensuremath{\IfNoValueTF{#1}
    {#3}%code when no optional argument is passed}
    {\IfNoValueTF{#2}
     {#3}
     {#3\left( #1 \| #2 \right)}
    }}%
}
\NewDocumentCommand{\kl}{o o}{
    \distance[#1][#2]{D_\text{KL}}
}

\DeclareMathOperator*{\argmax}{arg\,max}
\DeclareMathOperator*{\argmin}{arg\,min}

\newcommand{\inv}[1]{\frac{1}{#1}}

\newcommand{\card}[1]{\ensuremath{\left|#1\right|}}

\NewDocumentCommand{\norm}{m}{
 \ensuremath{\|#1\|}
}

\NewDocumentCommand{\ind}{o}{\ensuremath{
    \IfNoValueTF{#1}
        {\mathbbm{1}}
        {\mathbbm{1}\!\left[#1\right]}
    }
}

\newcommand{\expectation}[2]{\mathbb{E}_{#1}\left[#2\right]}
\newcommand{\variance}[2]{\mathbb{V}_{#1}\left[#2\right]}

\NewDocumentCommand{\entropy}{m}{\ensuremath{\mathbb{H} \left(#1\right)}}
\NewDocumentCommand{\gemini}{o o m}{\ensuremath{
    \IfNoValueTF{#1}
    {\IfNoValueTF{#2}
       {\I}
       {\I^{#2}}
    }
    {\IfNoValueTF{#2}
       {\I_{#1}}
       {\I_{#1}^{#2}}
    }
    \left(#3\right)
}}
\NewDocumentCommand{\mi}{m}{\ensuremath{\I \left(#1\right)}}

\begin{document}

\title{A tutorial on discriminative clustering and mutual information}

\date{}

\author[1]{Louis Ohl}
\author[2,3]{Pierre-Alexandre Mattei}
\author[2,4]{Frédéric Precioso}

\affil[1]{Linköping University, Division of Statistics and Machine Learning - Linköping, Sweden}
\affil[2]{Inria, Université Côte d'Azur, Maasai team - Nice, France}
\affil[3]{CNRS - LJAD}
\affil[4]{CNRS - I3S}

\maketitle

\begin{abstract}
To cluster data is to separate samples into distinctive groups that should ideally have some cohesive properties. Today, numerous clustering algorithms exist, and their differences lie essentially in what can be perceived as ``cohesive properties''. Therefore, hypotheses on the nature of clusters must be set: they can be either generative or discriminative. As the last decade witnessed the impressive growth of deep clustering methods that involve neural networks to handle high-dimensional data often in a discriminative manner; we concentrate mainly on the discriminative hypotheses. In this paper, our aim is to provide an accessible historical perspective on the evolution of discriminative clustering methods and notably how the nature of assumptions of the discriminative models changed over time: from decision boundaries to invariance critics. We notably highlight how mutual information has been a historical cornerstone of the progress of (deep) discriminative clustering methods. We also show some known limitations of mutual information and how discriminative clustering methods tried to circumvent those. We then discuss the challenges that discriminative clustering faces with respect to the selection of the number of clusters. Finally, we showcase these techniques using the dedicated Python package, GemClus, that we have developed for discriminative clustering.
\end{abstract}

\section{Introduction}

\emph{Clustering} is a fundamental learning task that involves separating data samples into several groups, each named cluster, without using labels. A cluster should be a cohesive set of elements that share some common properties. Ideally, these commonly shared properties should allow us to discriminate one cluster from the other. Therefore, this task is useful for exploring and uncovering knowledge in data analysis, \eg biology with microarray analysis~\citep{mclachlan_mixture_2002, sturn_genesis_2002}, customer segmentation~\citep{kashwan_customer_2013, kansal_customer_2018}, social network analysis~\citep{bedi_community_2016, himelboim_classifying_2017}, political campaign analysis~\citep{bode_candidate_2015} or disease phenogroup discovery~\citep{kwak_unsupervised_2020}.

Given a potentially large collection of \emph{data samples} $\x_i$ gathered into a dataset $\dataset$, a clustering algorithm is a model $f$ that assigns each sample $\x_i$ to a cluster $y_i$. While $\x_i$ may be from a high-dimensional space $\dataspace$ \eg images, tabular entries, or graphs, the cluster assignment $y$ is only an integer bounded by the maximum desired number of clusters $K$. Formally:
\begin{equation}\label{eq:intro_clustering_model}
    \begin{array}{r l}
        f: \dataspace &\rightarrow \naturalto{K}, \\
         \x_i &\mapsto f(\x_i)=y_i.
    \end{array}
\end{equation}

Conceptually, clustering belongs to the family of \emph{unsupervised learning}. This means that the dataset does not contain any information about a potential ideal target. Therefore, we do not have explicit information guiding the optimisation of the model $f$.

In the absence of such targets, clustering hinges on two main questions~\citep{hennig_what_2015}:
\begin{enumerate}
\item The first main question concerns the assessment of correct clustering. We are interested here in knowing if the discovered clusters are insightful and teach us something about the data. However, we must emphasise that there is no global consensus on the definition of a cluster. Indeed, it was shown that each clustering algorithm cannot satisfy simultaneously desirable properties~\citep{kleinberg_impossibility_2003}. In general, the use case drives the need, and the \textit{``clustering is in part in the eye of the beholder''}~\citep{estivill-castro_why_2002}. We may thus say that there are \emph{no absolute best clustering} algorithm, yet some may be relatively better depending on the context.
\item The second major question to address in clustering is the actual, sometimes called optimal, number of clusters \ie the value of $K$ in Eq.~(\ref{eq:intro_clustering_model}). For instance, a clustering algorithm could find more insightful clusters when searching only for 5 of them instead of 10. Perhaps more clusters would be more beneficial. Yet again, the lack of a formal definition of clusters implies that no method can be absolutely better than others in finding the correct number of clusters if that number is even existing. However, by restricting clusters to a narrow definition, there are methods to assess the quality of the number of clusters~\citep{tibshirani_estimating_2001, davies_cluster_1979, rousseeuw_silhouettes_1987,biernacki_assessing_2000}.
\end{enumerate}

During the last decade, clustering algorithms benefitted from the integration of neural networks \eg \citet{li_contrastive_2021, park_improving_2021, ronen_deepdpm_2022}. The main motivation is that \textit{``conventional clustering methods usually have poor performance on high-dimensional data, due to the inefficiency of similarity measures used in these methods''} \citep{min_survey_2018}. Therefore, the integration of neural networks in clustering, called deep clustering, is \textit{``expected to continue the tradition of clustering algorithms and to expand their ability to elucidate the hidden structure in big data [...]''} \citep{nutakki_introduction_2019} owing to their representation power. As proposals of deep clustering algorithm increase, taxonomy proposals have followed through surveys. We can start with \citet{min_survey_2018} who proposed to classify clustering algorithms depending on the nature of their neural network architecture: auto-encoder, Deep Neural Network (DNN), variational auto-encoder (VAE), or generative adversarial network (GAN). It is interesting that VAE and GANs are considered as different from DNNs because they are themselves built using DNNs. The initial motivation is that the DNN category only employs a single clustering loss. It is notable that such a taxonomy omits the nature of the clustering loss function, as \citet{min_survey_2018} argue that \textit{``since the essence of deep clustering is to learn a clustering-oriented representation, it is not suitable to classify methods according to the clustering loss''}. This taxonomy is still presented for the single-view clustering algorithms category in a survey by \citet{ren_deep_2024}.

Later reviews motivated that loss functions should be also part of the taxonomy, notably because a \textit{``good cost or loss function [...] captures what a good representation or clustering is''} \citep{nutakki_introduction_2019}. In this sense, \citet{nutakki_introduction_2019} classified deep clustering algorithms depending on their training procedure: sequential \ie one algorithm after another, joint, \ie simultaneously training the deep learning backbone and learning clusters, or closed-loop multistep clustering, \ie iteratively refining the training of multiple algorithms. Under slightly different names, this taxonomy is kept by \citet{zhou_comprehensive_2022}, with the addition of generative models as a different category. For an overview of current deep clustering algorithm, we recommend reading \citet{ren_deep_2024}, and for understanding the current definitions and limitations of deep clustering building blocks as well as current benchmark procedures, we recommend \citet{zhou_comprehensive_2022}.

Our goal in this tutorial is not to introduce a specific novel taxonomy. Instead, we propose to focus on discriminative clustering algorithms. These models are underlying some of the categories above: joint clustering, iterative clustering \cite{nutakki_introduction_2019}, and are often based on DNN architectures in the taxonomy of \citet{min_survey_2018}. Discriminative clustering is the opposite approach to generative clustering, which encompass GANs and VAEs. In line with the deep clustering hope of leveraging powerful representation, discriminative clustering models do not set parametric assumptions on the data distribution. Therefore, such deep discriminative methods seem to partially address (or omit) the major question of assessment (1): we hope to get a good representation for clustering, but do not question if the mean to obtain such a representation guarantees that they are insightful. When learning representations or clustering using discriminative models, assumptions remain and are hidden in other details, especially thee objective function. This is the case the case of the mutual information between data and clusters that has been a cornerstone of the development of discriminative clustering. In order to discuss the current state of such deep clustering methods, we propose to review the historical joint evolution of discriminative clustering and mutual information. We will draw parallels with generative models and classification methods for further understanding. The contributions of this paper are:
\begin{itemize}
    \item An accessible introduction to the field of discriminative clustering (Sections~\ref{sec:modelling}, \ref{sec:challenge} and \ref{sec:other})
    \item A historical account on the usage of mutual information for discriminative clustering leading to modern deep clustering methods. Notably, we present links between lines of research corresponding to different views on mutual information in discriminative clustering. (Section~\ref{sec:thrive})
    \item A highlight on the interpretation of clustering models in contrastive learning context. We show how the optimisation of a lower bound on MI implicitly minimises a variational inference problem between data and augmentations. (Section~\ref{sec:thrive})
    \item An emphasis on the call for model selection strategies that are in line with those novel clustering models, a topic that is often omitted. (Section~\ref{sec:model_selection})
    \item A showcase of the discussed methods using GemClus, a Python package that we developed. (Section~\ref{sec:example})
\end{itemize}

To the best of our knowledge, this tutorial is the first of a kind on discriminative clustering and the different lines of research that arose in this field. In related surveys \cite{jain_data_1999, min_survey_2018, nutakki_introduction_2019, zhou_comprehensive_2022, ren_deep_2024, wei_overview_2024}, the word discriminative appears to qualify learnt features, often in the sense of their ability to make a distinction between two samples.

We start by discussing in Section~\ref{sec:modelling} the two different types of modelling for clustering: generative and discriminative. We first detail the nature of the hypotheses in generative clustering, and existing solutions for major questions (1) and (2).

We continue with the challenges arising from the discriminative formulation of clustering in Section~\ref{sec:challenge} for optimising the parameters of the models. By drawing a parallel with classification tasks, we show how mutual information is introduced as a natural objective for learning the model parameters. We finish the section by highlighting the known limitations of mutual information. Then, we complete the presentation of objectives by detailing other clustering objectives, \eg K-means, in Section~\ref{sec:other}.

In Section~\ref{sec:thrive}, we dive in the history of mutual information as a clustering objective function. We start by discriminative models perceived as decision boundaries decoupled from the objective function~\citep{bridle_unsupervised_1992} and show how the progressive implicit need to regularise mutual information led to more intricate models where the model is no longer decoupled from the objective function, as in contrastive learning~\citep{li_contrastive_2021}.

We then draw our attention in Section~\ref{sec:model_selection} to choosing a number of clusters. We notably insist that the selection strategy should be aligned with the clustering hypotheses set by the dual combination model and objective.

We finish in Section~\ref{sec:example} by presenting an example of discriminative clustering application on a simple dataset. We notably present here \emph{GemClus}, a package that covers discriminative clustering methods presented in Section~\ref{sec:thrive}.

The notations used throughout the paper are listed in Table~\ref{tab:notations}.
\begin{table}[t]
    \centering
    \caption{Notations used in this paper.}
    \label{tab:notations}
    \begin{tabular}{p{0.1\linewidth} p{0.35\linewidth} p{0.1\linewidth} p{0.35\linewidth}}
    \toprule
    Notation & Description & Notation & Description\\
    \midrule
    $n$ & Number of observations/samples&
    $d$ & Number of features per sample\\
    $K$ & Number of clusters&
    $\Delta^{K-1}$ & The $K-1$-simplex\\
    $\dataspace$ & Data space&
    $\x$ & An observation from $\dataspace$\\
    $\x_{ij}$ & The $j$-th feature of the $i$-th observation in the dataset $\dataset$&
    $\naturalto{n}$ & The set of natural integers from 1 to $n$\\
    $\cluster$ & Cluster containing samples&
    $y$ & The cluster membership\\
    $\mathcal{Z}$ & An intermediate representation space&
    $\vec{z}$ & A continuous variable from $\mathcal{Z}$\\
    $\mathcal{H}$ & Reproducing kernel Hilbert space&
    $\card{\cdot}$ & Cardinal\\
    $\p$ & Distribution with parameters $\theta$&
    $\entropy{\cdot}$ & Entropy of a random variable\\
    $\expectation{p}{\cdot}$ & Expectation of a random variable under distribution $p$&
    $\variance{p}{\cdot}$ & Variance of a random variable under distribution $p$\\
    $\normal$ & Multivariate and univariate Gaussian distribution&
    $\delta_{\x}$ & Delta Dirac distribution located in position $\x$\\
    $\ind$ & Indicator function&
    $\I$ & Mutual information\\
    $\objective$&An objective function&$E$&Energy of an energy-based model\\
    $\norm{\cdot}$ & Norm&
    $\kl[\cdot][\cdot]$ & Kullback-Leibler divergence\\
    $c(\cdot,\cdot)$ & A distance function (or cost) between samples in $\dataspace$&
    $\kappa(\cdot, \cdot)$ & A kernel function between samples in $\dataspace$\\
    \bottomrule
    \end{tabular}\vspace{-1.5\baselineskip}
\end{table}
\section{Modelling frameworks}
\label{sec:modelling}

Clustering is the task of grouping data samples. Each data sample is formally described as a random variable $\x$, taking values from $\dataspace$. This variable is uni- or multidimensional, with a mix of continuous and/or discrete dimensions. It also possible that $\dataspace$ concerns graphs. We consider that we have a dataset of $n$ independent and identically distributed samples $\dataset = \{\x_i\}_{i=1}^n$. The membership of the cluster is indicated by $y$, a discrete random variable taking values in $\naturalto{K}$, where $K$ is the number of clusters to determine.

To link these two random variables, a model is required. However, the nature of the assumptions made for the model greatly impacts the algorithmic procedures for learning. Here we describe two major contrasting frameworks: the \emph{generative} and the \emph{discriminative} models.

\subsection{Starting from Bayes theorem}
\label{ssec:modelling_bayes}

A clustering algorithm can be described as a probability distribution $p$ that assigns a cluster membership $y$ to  a given sample $\x$. This distribution is controlled by a set of parameters $\theta$. Thus, a clustering model is the parameterisation of the distribution $p$ by $\theta$ and so it is the conditional probability $\p(y\mid \x)$. According to the Bayes theorem, we can devise a definition of the clustering model $\p(y\mid \x)$:
\begin{equation}\label{eq:bayes_theorem}
    \p(y\mid \x) = \frac{\p(\x\mid y)\p(y)}{\p(\x)}.
\end{equation}

This theorem highlights that building a distribution binding clusters $y$ and data $\x$ necessarily implies the existence of three other distributions: one for the generation of the data given a cluster $\p(\x\mid y)$, one for the proportion of clusters $\p(y)$ and one for the probability of observing the data $\p(\x)$, often called \emph{likelihood}. We further note that as the clustering distribution is dependent on $\theta$, other distributions are consequently also dependent on $\theta$, unless specified otherwise.

Given a clustering model $\p(y\mid \x)$, the final clustering of a dataset is the assignment of each sample in the dataset $\dataset$ to the cluster for which the conditional probability is maximal. The $k$-th cluster is defined as:
\begin{equation}
    \cluster_k = \{\x \in \dataset \mid k = \argmax_{y} \p(y\mid \x)\}.
\end{equation}

The definition offered by the Bayes theorem on the clustering models brings two different ways of defining the clustering model and its parameters. The first one is the generative modelling which views the clustering membership as a latent variable explaining how the data was generated, and the second one is the discriminative modelling which seeks immediately the clusters from the data. With probabilistic graphical models~\citep[Chapter 1]{koller_probabilistic_2009}, we can summarise both views with Figure~\ref{fig:modellings}.

\subsection{Generative models}
\label{ssec:modelling_generative}

\subsubsection{Definition}

\begin{figure}
    \centering
    \subfloat[Generative modelling]{
        %\begin{scaletikzpicturetowidth}{0.45\textwidth}
        \hspace{0.1\linewidth}
        \begin{tikzpicture}%[scale=\tikzscale]
        	\node[shape=circle,draw=black] (y) at (0,0) {$y$};
        	\node[shape=circle,draw=black, right=of y, fill=lightgray] (x) {$\x$};
        	\draw[->,very thick] (y) --  (x);
        \end{tikzpicture}
        %\end{scaletikzpicturetowidth}
        \label{sfig:generative}
        \hspace{0.1\linewidth}
    }
    \subfloat[Discriminative modelling]{
        \hspace{0.1\linewidth}
        %\begin{scaletikzpicturetowidth}{0.3\textwidth}
        \begin{tikzpicture}%[scale=\tikzscale]
        	\node[shape=circle,draw=black,fill=lightgray] (x) at (0,0) {$\x$};
        	\node[shape=circle,draw=black, right=of x] (y) {$y$};
        	\draw[->,very thick] (x) --  (y);
        \end{tikzpicture}
        %\end{scaletikzpicturetowidth}
        \label{sfig:discriminative}
        \hspace{0.1\linewidth}
    }
    \caption[The generative and discriminative modelling frameworks]{The generative and discriminative modelling frameworks for clustering models. Observed variables are shaded.}
    \label{fig:modellings}
\end{figure}
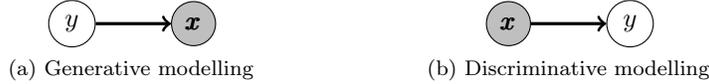

In generative modelling, knowing the latent cluster $y$ is sufficient to describe the distribution of its associated data $\x$, see Figure~\ref{sfig:generative}. The design of this model is therefore focused on the right hand-side factors of the Bayes theorem in Eq.~(\ref{eq:bayes_theorem}): a generative model is the design of $\p(\x\mid y)$, with learnable proportions $\p(y)$~\citep{bouveyron_model-based_2019}. Often, the distribution $\p(\x\mid y)$ is very simple, \eg Gaussian distribution. The generative approach can be interpreted as creating a ready-made template of how clusters would look like, then stretching the template until it fits as best as possible the observed data. The central idea of fitness is often measured with the likelihood:
\begin{equation}
    \p(\x) = \sum_{k=1}^K \p(\x\mid y=k)\p(y=k).
\end{equation}

The intuition is that a model similar to the true process that generated the data should be likely to generate similar samples to those observed. The clustering distribution $\p(y\mid \x)$ is then a consequence of the modelling. In fact, a generative model indirectly specifies the clustering distribution because it is implicitly proportional to the generative process $\p(\x\mid y)\p(y)$.

\begin{figure}
    \centering
    \pgfmathdeclarefunction{gauss}{2}{%
      \pgfmathparse{1/(#2*sqrt(2*pi))*exp(-((x-#1)^2)/(2*#2^2))}%
    }
    \subfloat[25\% of red cluster]{
        \begin{tikzpicture}
            \begin{axis}[
              no markers, domain=-5:5, samples=100,
              axis lines*=left, xlabel=$x$,
              every axis y label/.style={at=(current axis.above origin),anchor=south},
              every axis x label/.style={at=(current axis.right of origin),anchor=west},
              height=5cm, width=12cm,
              xtick={-2,2}, ytick=\empty,
              enlargelimits=false, clip=false, axis on top,
              grid = major,
              width=0.3\linewidth,
              height=0.1\paperheight
              ]
              \addplot [very thick,red] {0.25*gauss(-2,1)};
              \addplot [very thick,blue] {0.75*gauss(2,1)};
            \end{axis}
        \end{tikzpicture}
    }
    \subfloat[50\% of red cluster]{
        \begin{tikzpicture}
            \begin{axis}[
              no markers, domain=-5:5, samples=100,
              axis lines*=left, xlabel=$x$,
              every axis y label/.style={at=(current axis.above origin),anchor=south},
              every axis x label/.style={at=(current axis.right of origin),anchor=west},
              height=5cm, width=12cm,
              xtick={-2,2}, ytick=\empty,
              enlargelimits=false, clip=false, axis on top,
              grid = major,
              width=0.3\linewidth,
              height=0.1\paperheight,
              ]
              \addplot [very thick,red] {0.5*gauss(-2,1)};
              \addplot [very thick,blue] {0.5*gauss(2,1)};
            \end{axis}
        \end{tikzpicture}
    }
    \subfloat[90\% of red cluster]{
        \begin{tikzpicture}
            \begin{axis}[
              no markers, domain=-5:5, samples=100,
              axis lines*=left, xlabel=$x$,
              every axis y label/.style={at=(current axis.above origin),anchor=south},
              every axis x label/.style={at=(current axis.right of origin),anchor=west},
              height=5cm, width=12cm,
              xtick={-2,2}, ytick=\empty,
              enlargelimits=false, clip=false, axis on top,
              grid = major,
              width=0.3\linewidth,
              height=0.1\paperheight
              ]
              \addplot [very thick,red] {0.9*gauss(-2,1)};
              \addplot [very thick,blue] {0.1*gauss(2,1)};
            \end{axis}
        \end{tikzpicture}
    }
    \caption[Example of one-dimensional Gaussian mixture]{An example of generative models with 2 clusters in 1 dimension. The parameters $\theta$ of the model comprise the Gaussian distribution locations $\vec{\mu}_\text{red}=-1$, $\vec{\mu}_\text{blue}=2$, the scales $\sigma_\text{red}^2=\sigma^2_\text{blue}=1$ and the proportions of each clusters.}
    \label{fig:gaussian_mixture}
\end{figure}
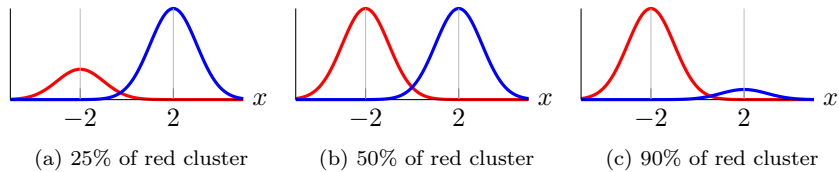

We give a simple example of a generative model where each cluster obeys a Gaussian distribution in Figure~\ref{fig:gaussian_mixture}. Therefore, each cluster distribution is written $\p(\x\mid y) = \mathcal{N}(\x\mid\vec{\mu}_y,\sigma^2_y)$ and each cluster proportion $\p(y)=\pi_y$. The parameters $\theta$ comprise the location $\vec{\mu}_y$, the scale $\sigma^2_y$ and proportion $\pi_y$ for each cluster. In the one-dimensional binary example from Figure~\ref{fig:gaussian_mixture}, we have a total of 6 parameters with $\theta=\{\vec{\mu}_\text{red}, \sigma^2_\text{red}, \pi_\text{red}, \vec{\mu}_\text{blue}, \sigma^2_\text{blue}, \pi_\text{blue}\}$. The resulting likelihood is defined as:
\begin{align}
    \p(\x) &= \p(y=\text{red}) \p(\x \mid y=\text{red}) + \p(y=\text{blue}) \p (\x \mid y=\text{blue})\\
    &=\pi_\text{red} \mathcal{N}(\x \mid \vec{\mu}_\text{red}, \sigma^2_\text{red}) + \pi_\text{blue} \mathcal{N}(\x \mid \vec{\mu}_\text{blue}, \sigma^2_\text{blue}).
\end{align}

For a mixture of 2 $d$-dimensional Gaussian distribution with proportions $\pi_1$, $\pi_2$, locations $\vec{\mu}_1$, $\vec{\mu}_2$ and the same covariance $\vec{\Sigma}$, it is possible to show that the clustering distribution is defined by~\citep[Eq. 4.64]{bishop_pattern_2007}:
\begin{equation}\label{eq:gaussian_boundary}
    \p(y = 1\mid \x) = \text{Sigmoid}(\vec{w}^\top \x + b),
\end{equation}

\noindent where the sigmoid function is defined for all real values $a$:
\begin{equation}
    \text{Sigmoid}(a) = \frac{1}{1-e^{-a}},
\end{equation}

\noindent and the coefficients are:
\begin{equation}
    \vec{w} = \vec{\Sigma}^{-1}(\vec{\mu}_1-\vec{\mu}_2),
\end{equation}

\noindent and:
\begin{equation}
    b = -\inv{2}\vec{\mu}_1^\top\vec{\Sigma}^{-1}\vec{\mu}_1 + \inv{2}\vec{\mu}_2^\top\vec{\Sigma}^{-1}\vec{\mu}_2 + \log \frac{\pi_1}{\pi_2}.
\end{equation}

We observe in Eq.~(\ref{eq:gaussian_boundary}) that the clustering distribution $\p(y\mid \x)$ is therefore drawn around a linear decision boundary of coefficients $\vec{w}$ and $b$ when the covariances are equal between 2 Gaussian distributions.

\subsubsection{Learning}

To optimise the parameters, the most straightforward method is \emph{maximum likelihood}, \ie maximising the value of the likelihood $\p(\x)$ over all samples $\x_i$ in the dataset $\dataset$. Under the common assumption of i.i.d. samples in the dataset, the maximum likelihood parameter is defined as:
\begin{equation}
    \hat{\theta}_\text{MLE} \in \argmax_\theta \prod_{\x\in\dataset} \p(\x).
\end{equation}

However, this likelihood is often challenging to maximise, due to the unobserved and latent nature of the cluster membership $y$. In such a case, an Expectation-Maximisation (EM) algorithm can be used to find a local maximum for the likelihood~\citep{mclachlan_em_2007}. This algorithm alternates between two steps. In the first step, called Expectation, the probability of cluster membership is estimated using the current state of the model parameters $\theta$. In the second step, called Maximisation, the parameters $\theta$ are optimised to maximise the likelihood given the current cluster membership probabilities. This procedure is repeated multiple times until convergence. Still today, this algorithm remains one of the most commonly used in mixture modelling.

The most common example of a generative clustering model is the Gaussian Mixture Model (GMM), where each cluster is assumed to follow a Gaussian distribution~\citep{banfield_model-based_1993, bouveyron_model-based_2019}. Other examples comprise mixtures Gaussian copulae for handling mixed types of variables~\citep{marbac_model-based_2017}, mixtures of multivariate $t$-distribution~\citep{peel_robust_2000} for robustness against outliers thanks to heavily tailed distributions, for instance, applied on Box-Cox transformed data~\citep{lo_automated_2008}, mixtures of Poisson distributions~\citep{karlis_exact_2008}, or mixtures of multinomial distributions~\citep{goodman_exploratory_1974}.

\subsubsection{Selecting a number of clusters}

Model-based clustering benefits from a large panel of statistical tools. In particular, it is possible to assess the adequation of the model to the data and validate a correct choice for the number of clusters with internal scores. For example, the Bayesian Information Criterion (BIC, ~\citealp{schwarz_estimating_1978}) is defined for the parameters $\hat{\theta}_\text{MLE}$ obtained after training:
\begin{equation}
    \text{BIC}(K) = 2 \sum_{\x\in\dataset} \log p_{\hat{\theta}_\text{MLE}}(\x) - \nu_K \log{n},
\end{equation}
where $\nu_K$ is the number of free parameters in $\hat{\theta}_\text{MLE}$ for $K$ clusters. For example: in the previous GMM example, we had $\nu_K=5$ because the proportions of the clusters $\pi_\text{red}$ and $\pi_\text{blue}$ depend on each other: $\pi_\text{blue} = 1 -\pi_\text{red}$. BIC encourages models to fit well the data with a strong likelihood while maintaining a low number of parameters. This criterion focuses on the good number of components that fit well the data in terms of likelihood. However, this does not imply that it is correct for clustering the data. A more adequate internal score for mixture-based clustering models is the Integrated Complete Likelihood (ICL, \citealp{biernacki_assessing_2000}) for $K$ clusters and $M$ components:
\begin{equation}
    \text{ICL}(K,M) = \text{BIC}(M) - \sum_{\x\in\dataset} \sum_{y=1}^K p_{\hat{\theta}_\text{MLE}}(y\mid \x) \log p_{\hat{\theta}_\text{MLE}}(y\mid \x).
\end{equation}

The ICL penalises the BIC by subtracting the entropy of the cluster memberships of the model. As the entropy becomes smaller when the model decisions are clear-cut, ICL encourages models for which the number of clusters leads to distinct separations.

\subsubsection{Generative hypotheses}

The difficulty of generative models for clustering lies essentially in the choice of the generative distribution $\p(\x\mid y)$. In practice, the choice of distribution, \eg Poisson, Gaussian, is often guided by the expertise developed on the dataset. However, the statistical challenges emerging from the high-dimensional nature of modern data call for adequate regularisations and constraints over the distributions~\citep[Chapter 8]{bouveyron_model-based_2019}.

Beyond the usage of well-known distributions with scalar or matrix parameters, the last decade witnessed the rise of generative modelling with neural networks, notably with the Variational Auto-Encoders (VAE, \citealp{kingma_adam_2014, rezende_stochastic_2014}) and the Generative Adversarial Networks (GAN, \citealp{goodfellow_generative_2014}). Both methods aim to build generative models in which the latent variable $\z$ is \emph{continuous} contrary to generative clustering models.
Starting from easy-to-sample distributions, these models transform low-dimensional latent variables into high-dimensional variables using complex non-linear transformations.
VAEs, as part of the broader class of Deep Latent Variable Models (DLVM), parameterise the generative distribution with the output of a function $g$ from the latent variable: $p(x\mid g(\z))$. For example, $g$ can be a neural network that returns the mean and diagonal covariance of a Gaussian distribution. To optimise the generative process, an amortised variational inference method is often used with an encoder network as the proposal distribution for approximating the posterior distribution of the latent variable, hence the name variational auto-encoder. GANs focus instead on optimising a zero-sum game, where the generative distribution must produce samples of sufficient quality to fool a discriminative network, sometimes called a critic. The goal of this discriminator is to differentiate true samples coming from a target distribution (defined by the data), from samples created \emph{de novo} by the generator. The core limitation of these initial models in the context of clustering is the usage of a continuous latent variable, the code, rather than a discrete variable, the cluster. There are proposals to make the latent variable categorical~\citep{jang_categorical_2017}, but this does not imply that they were intended for clustering. Thus, extensions were proposed to adapt VAEs and GANs for clustering \citep{jiang_variational_2017, dilokthanakul_deep_2016, springenberg_unsupervised_2015, mello_top-down_2022}.

In general, generative modelling presents compelling tools for clustering. However, the difficulty in making assumptions on the data distribution is one of the main limitations to the success of the model. Moreover, in the context of mixed types of variables, this choice of parametric assumption becomes even harder. Often, assumptions of independence between categorical and continuous variables are made to achieve clustering with mixture models, of mixed-type variables~\citep{marbac_model-based_2017, marbac_variable_2017, ma_vaem_2020}.

\subsection{Discriminative models}
\label{ssec:modelling_discriminative}

To avoid the burden of choosing parametric assumptions following the generative modelling, we turn to the discriminative point of view. In this context, we do not take any assumption at all regarding the data distribution and denote it $\pdata(\x)$. However, we assume to be able to sample from a given dataset $\dataset$. The discriminative framework can be interpreted as taking the data as is, and inferring clusters instead of finding assignments suiting a notion of likelihood. Thus, we only design a discriminative model $\p(y\mid \x)$ and we obtain:
\begin{equation}
    \p(\x,y) = \pdata(\x) \p(y\mid \x).
\end{equation}

From a generative perspective, the discriminative modelling corresponds to the joint distribution of a data distribution and clustering model with decoupled parameters~\citep{minka_discriminative_2005}. The data is generated by a set of external parameters $\theta^\prime$ and the complete model is written $p_{\theta,\theta^\prime}(\x,y) = \p(y\mid \x)p_{\theta^\prime}(\x)$. In contrast to this generative view, we do not even assume the existence of such parameters in our discriminative models when writing $\pdata(\x)$.

In this paper, we use the term \emph{discriminative} in the sense, for instance, of \citet{minka_discriminative_2005} and \citet{bishop_pattern_2007}. Note that some other clustering methods employ the term \emph{discriminative} to designate the finding of discriminative subspaces, \eg using Fisher discriminant analysis, linear discriminant analysis~\citep{torre_discriminative_2006,ye_discriminative_2007,bouveyron_simultaneous_2012}.

This framework can be used conveniently with any function $\psi_\theta$ whose outputs lie in the $K-1$ simplex, where $K$ is the number of clusters. This output can then be considered as the parameters of a categorical distribution defining the conditional cluster membership:
\begin{equation}
y\mid \x \sim \text{Categorical}(\psi_\theta(\x)).
\end{equation}

Thanks to the degrees of freedom for the definition of $\psi_\theta$, the discriminative clustering framework can tolerate various softmax-ended neural networks. For example, we can consider logistic regressions where the discriminative model takes the form:
\begin{equation}\label{eq:logistic_regression}
    \psi_\theta(\x) = \text{Softmax}(\vec{W}^\top\x + \vec{b}),
\end{equation}

\noindent where the parameters are $\theta=\{\vec{W}, \vec{b}\}$. The softmax function is defined for any real vector and returns a stochastic vector:
\begin{equation}
    \text{Softmax}(\z) = \left[ \frac{e^{\z_1}}{Z}, \ldots, \frac{e^{\z_i}}{Z}, \ldots, \frac{e^{\z_d}}{Z} \right]^\top, \quad\text{with}\quad Z = \sum_{i=1}^d e^{\z_i}.
\end{equation}

Thus, in contrast to generative modelling, discriminative modelling encompasses our hypotheses through the design of the decision boundary. For example, Eq.~(\ref{eq:logistic_regression}) shows that a logistic regression is the discriminative equivalent of a mixture of Gaussian distributions with equal variances: both their decision boundaries are linear, as we showed in Eq.~(\ref{eq:gaussian_boundary}). 

Beyond the scope of clustering, discriminative modelling is the core approach used for classification with neural networks. Thus, a neural network that was well designed for a classification task on a specific type of data can be used immediately for clustering. However, the discriminative perspective in the context of clustering severely affects the procedures to learn optimal parameters. 

\section{The challenge of learning in discriminative clustering}
\label{sec:challenge}

We now discuss how discriminative models can be trained and the implied properties.

\subsection{The shortcomings of classical statistical tools}
\label{ssec:challenge_shortcomings}

Due to the absence of a model on the data distribution $\pdata(\x)$, we cannot use maximum likelihood to learn the optimal parameters, and consequently neither expectation-maximisation nor variational inference. Additionally, we cannot construct a generative cluster distribution $\p(\x\mid y)$, because we we do not have access to the distribution $\pdata(\x)$. In other words, a discriminative model cannot generate samples outside the dataset. One quantity we can actually estimate is the vector of proportions of the clusters through marginalisation:
\begin{equation}\label{eq:discriminative_proportions}
    \p(y) = \expectation{\x \sim \pdata(\x)}{\p(y\mid \x)}.
\end{equation}

In practice, we cannot evaluate the density $\pdata(\x)$. This means that the value of the cluster proportions is only estimated by Monte Carlo on the dataset:
\begin{equation}
    \p(y) \approx \inv{\card{\dataset}} \sum_{\x \in \dataset} \p(y\mid \x).
\end{equation}

In summary, the absence of parametric assumptions on the data distribution leads to the absence of joint modelling:
\begin{equation}
    \underbrace{\p(y\mid \x)}_\text{Known} \times \underbrace{\pdata(\x)}_\text{\emph{Unknown}} = \underbrace{\p(\x\mid y)}_\text{\emph{Unknown}} \times \underbrace{\p(y)}_\text{Estimable} = \underbrace{\p(\x,y)}_\text{\emph{Unknown}}.
\end{equation}

The only thing we assume to be able to do is sampling from the data distribution, owing to the presence of a dataset $\dataset = \{\x_i\}_{i=1}^n$, which we use to estimate Eq.~(\ref{eq:discriminative_proportions}).

Therefore, a different approach must be taken for training discriminative models in clustering. To build such an approach, we will draw inspiration from the supervised models in classification that are often discriminative. These models leverage learning with an objective function that is decoupled from the model.

\subsection{Objective functions in classification}
\label{ssec:challenge_classification}

In the absence of parametric assumptions on the data, we are interested only in grouping the samples based on some meaningful criterion. In classification tasks, in contrast to clustering, we have access to labels that provide us with this guiding criterion. Therefore, we know that there exists an ideal distribution $\pdata(y\mid \x)$ that we must match as best as possible with our model $\p(y\mid \x)$. To that end, we must measure the distance of our current model from this ideal distribution and make it as close as possible. The most common distance between two distributions is the Kullback-Leibler (KL) divergence. For two arbitrary distributions $q_1$ and $q_2$, the KL divergence is defined as:
\begin{equation}
    \kl[q_1(\z))][q_2(\z)] = \expectation{\z \sim q_1(\z)}{\log \frac{q_1(\z)}{q_2(\z)}}.
\end{equation}

Note that the KL is in fact not a distance since it does not respect symmetry. Incorporating in this divergence the definition of our target model $\pdata(y\mid \x)$ and our classification model $\p(y\mid \x)$ would only tell us how far apart these two distributions are for a specific datum $\x$. Therefore, we wrap up this divergence in an expectation over the data distribution to ensure that \emph{on average}, our model sticks as best as possible to the targets. This data distribution is part of the ideal model $\pdata(\x)$. Thus, we obtain the average distance of our parameters to the data distribution $\objective(\theta)$:
\begin{align}\label{eq:kl_crosseentropy}
    \objective(\theta)&=\expectation{\x \sim \pdata(\x)}{\kl[\pdata(y\mid \x)][\p(y\mid \x)]}, \\ &=\expectation{\x \sim \pdata(\x), y\sim \pdata(y\mid \x)}{\log \frac{\pdata(y\mid \x)}{\p(y\mid \x)}},\\
    &= \expectation{\x\sim \pdata(\x), y\sim \pdata(y\mid \x)}{\log \pdata(y\mid \x)} - \expectation{\x \sim \pdata(\x), y\sim \pdata(y\mid \x)}{\log \p(y\mid \x)},\\
    &= \text{constant} - \expectation{\x \sim \pdata(\x)}{ \sum_{k=1}^K \pdata(y=k\mid \x) \log \p(y=k\mid \x)}.\label{eq:pseudo_crossentropy}
\end{align}

The constant here depends only on the ideal distribution and thus is not impacted by our choice of discriminative parameters $\theta$. Under the assumption that we have access to a dataset containing $n$ samples independently and identically distributed from $\pdata(\x, y)$, we can estimate the expectation using Monte Carlo. Noting the dataset $\dataset = \{(\x_i, y_i)\}_{i=1}^n$ which specifies the class to which each sample should belong, the target distribution can be approximated using a delta Dirac distribution for all $\x_i$: $\pdata(y\mid \x=x_i) \approx \ind[y=y_i]$. As we ignore the constant first term from Eq.~(\ref{eq:pseudo_crossentropy}), we get:
\begin{align}
    \objective(\theta) &= - \expectation{\x \sim \pdata(\x)}{ \sum_{k=1}^K \p(y=k\mid \x) \log \p(y=k\mid \x)}, \\
    &\approx - \inv{\card{\dataset}}\sum_{\x_i, y_i \in \dataset} \sum_{k=1}^K \ind[y_i = k] \log{\p(y=k\mid \x=\x_i)},\\
    &= -\inv{\card{\dataset}} \sum_{\x_i, y_i \in \dataset} \log{\p(y=y_i\mid \x=\x_i)}.
\end{align}

This objective is called \emph{cross-entropy} and is one of the most natural \emph{objective functions} (sometimes called \emph{costs}), in classification tasks and enjoys good properties: it is convex w.r.t. to the outputs of the model and ensures the minimisation of the KL divergence to the empirical estimate of $\pdata$. Moreover, minimising the cross-entropy is equivalent to maximising the likelihood of the data distribution in classification.

Owing to its differentiability and its sole dependence on the output of the model, this objective can be optimised by gradient descent under the condition that the model's outputs are differentiable w.r.t. $\theta$. This means differentiating the cross-entropy and backpropagating the derivative throughout the complete model parameters such that the cross-entropy gets lowered. When the derivative of the objective is 0, the model found a local minimum.

Throughout this example, we see that the notion of objective function depending on the output of the model providing a convenient gradient is an efficient solution for training discriminative models. The cross-entropy naturally emerged from the distance comparison between our model and a target model, thus guiding the gradient descent. We are interested in finding such an objective function that would fit into the context of clustering for discriminative models.

\subsection{Mutual information as a promising objective}
\label{ssec:challenge_mutual_information}

For the classification task, we relied on the distance in the KL sense to a target distribution to train the model. This led us to the cross-entropy loss. For the clustering task with generative modelling, we relied on the likelihood of the data describing how our model fits the data. Therefore, we must focus on a key property that is desirable in the discriminative clustering case. That key property is that the final clustering should reflect insights on the data distribution. Indeed, clusters inform us about the data and conversely, knowing the data informs us about the cluster. This notion is conveyed through the \emph{dependence} between two random variables. In our specific case: the clusters $y$ and the data $\x$ must be as dependent as possible.

To seek dependence between two random variables $\vec{\alpha}$ and $\vec{\beta}$, there exists a score that we can use as an objective function based on the KL divergence as well: \emph{Mutual Information} (MI), defined as:
\begin{equation}\label{eq:definition_joint_mutual_information}
    \mi{\vec{\alpha};\vec{\beta}} = \kl[p(\vec{\alpha}, \vec{\beta})][p(\vec{\alpha})p(\vec{\beta})].
\end{equation}

Mutual information can be seen as a measure of how dependent two random variables are: the greater, the more dependent. In fact, following Theorem~\ref{theorem:mi_independence}, a null mutual information value indicates that the clusters $y$ and the data $\x$ are independent, \ie unrelated.
\begin{theorem}[Independence in mutual information]\label{theorem:mi_independence}
Let $\vec{\alpha}$ and $\vec{\beta}$ be two random variables. Both variables are independent if and only if their mutual information is equal to 0.
\end{theorem}

\begin{remark}
Theorem~\ref{theorem:mi_independence} is a direct consequence of $\kl[p][q] = 0 \iff p=q$.
\end{remark}

Maximising mutual information to increase dependence between data and clusters seems a coherent objective. The upper bound of mutual information is the minimum of the entropies of $y$ and $\x$.

The definition from Eq.~(\ref{eq:definition_joint_mutual_information}) cannot be optimised as such because the joint model $\p(\x,y)$ and the data distribution $\pdata(\x)$ are unknown in the discriminative context. Fortunately, well-known properties of MI can invert the distributions on which the KL divergence is computed~\citep{bridle_unsupervised_1992,krause_discriminative_2010} via the product rule:
\begin{align}
    \mi{\x;y} &= \kl[\p(\x,y)][\pdata(\x)\p(y)],\\
    &= \expectation{\x,y\sim \p(x,y)}{\log\frac{\p(\x,y)}{\pdata(\x)\p(y)}},\\
    &= \expectation{\x\sim \pdata(x)}{\expectation{y\sim \p(y\mid \x)}{\log \frac{\p(y\mid \x)\pdata(\x)}{\pdata(\x)\p(y)}}}.
\end{align}

Then, we can simplify the factors on the data distribution inside the log and reidentify the KL divergence between a conditional distribution and a single marginal within an expectation:
\begin{equation} \label{eq:tractable_discriminative_mi}
    \mi{\x;y} = \expectation{\x \sim \pdata(\x)}{\kl[\p(y\mid \x)][\p(y)]}.
\end{equation}

Owing to the expectation on the data distribution, we can derive estimates of mutual information and its gradients w.r.t. $\theta$ using Monte Carlo. Thus, the usage of the product rule within the KL divergence to obtain an estimate is the key property to compute mutual information in discriminative modelling. This estimate, depends only on the output of the model $\p(y\mid \x)$ from which we can estimate the proportions of the clusters through marginalisation, recalling Eq.~(\ref{eq:discriminative_proportions}). Note that the cost function for clustering in Eq.~(\ref{eq:tractable_discriminative_mi}) is interestingly similar to the KL we minimised in supervised learning from Eq.~(\ref{eq:kl_crosseentropy}). In both equations, we sample on a data distribution and optimise the KL divergence between the model's clustering outputs and a target. In Eq.~(\ref{eq:kl_crosseentropy}), this target to reach is the ideal classifier on the dataset, and we minimise the KL. In Eq.~(\ref{eq:tractable_discriminative_mi}), the sampling distribution is the empirical approximation of the data distribution, the target is the cluster proportions, and we maximise the KL. Finally, we can maximise mutual information through gradient \emph{ascent}.

However, we must remember that we do not maximise the true mutual information. Indeed, we previously showed with Eq.~(\ref{eq:discriminative_proportions}) that the proportions $\p(y)$ are estimated with Monte Carlo. This means that the true proportions of the clusters are often not known in practice. 
Consequently, by rewriting this estimate as a proposal distribution $q(y)$ for the proportions of the cluster, we can unfold mutual information as follows~\citep{poole_variational_2019}:
\begin{align}
    \mi{\x;y} &= \expectation{\x\sim\pdata(\x)}{\kl[\p(y\mid \x)][\p(y) \times \frac{q(y)}{q(y)}]},\\
    &= \expectation{\x\sim \pdata(\x)}{\kl[\p(y\mid \x)][q(y)]} - \kl[\p(y)][q(y)],\\
    & \leq \expectation{\x \sim \pdata(\x)}{\kl[\p(y\mid \x)][q(y)]}.
\end{align}

This implies that we maximise only an upper bound of MI, even though the KL divergence between the estimated cluster proportions and the true proportions may be empirically negligible due to our unbiased estimate.

\begin{figure}[t]
    \begin{center}
    \subfloat[Unclustered samples]{
        \includegraphics[width=0.3\linewidth]{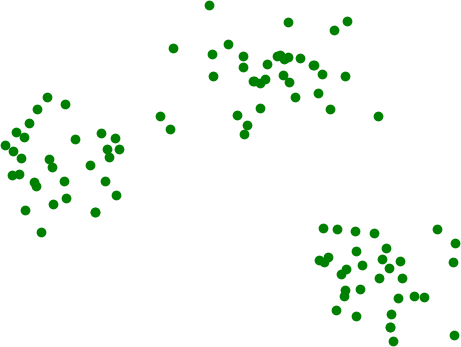}
        \label{sfig:nonparametric_unclustered}
    }\hfill
    \subfloat[Nonparametric model]{
        \includegraphics[width=0.3\linewidth]{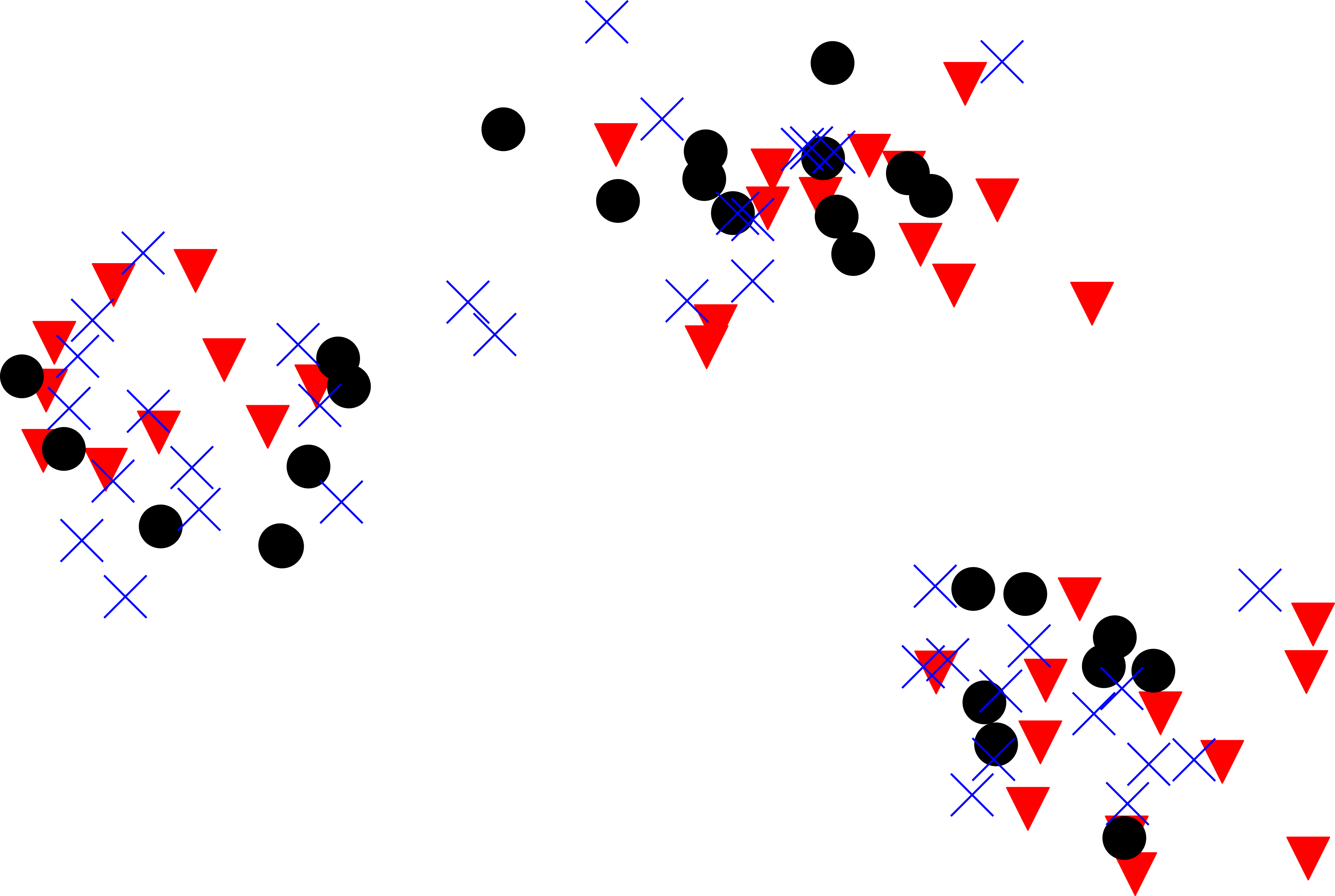}
        \label{sfig:nonparametric_clustered}
    }\hfill
    \subfloat[GEMINI solution]{
        \includegraphics[width=0.3\linewidth]{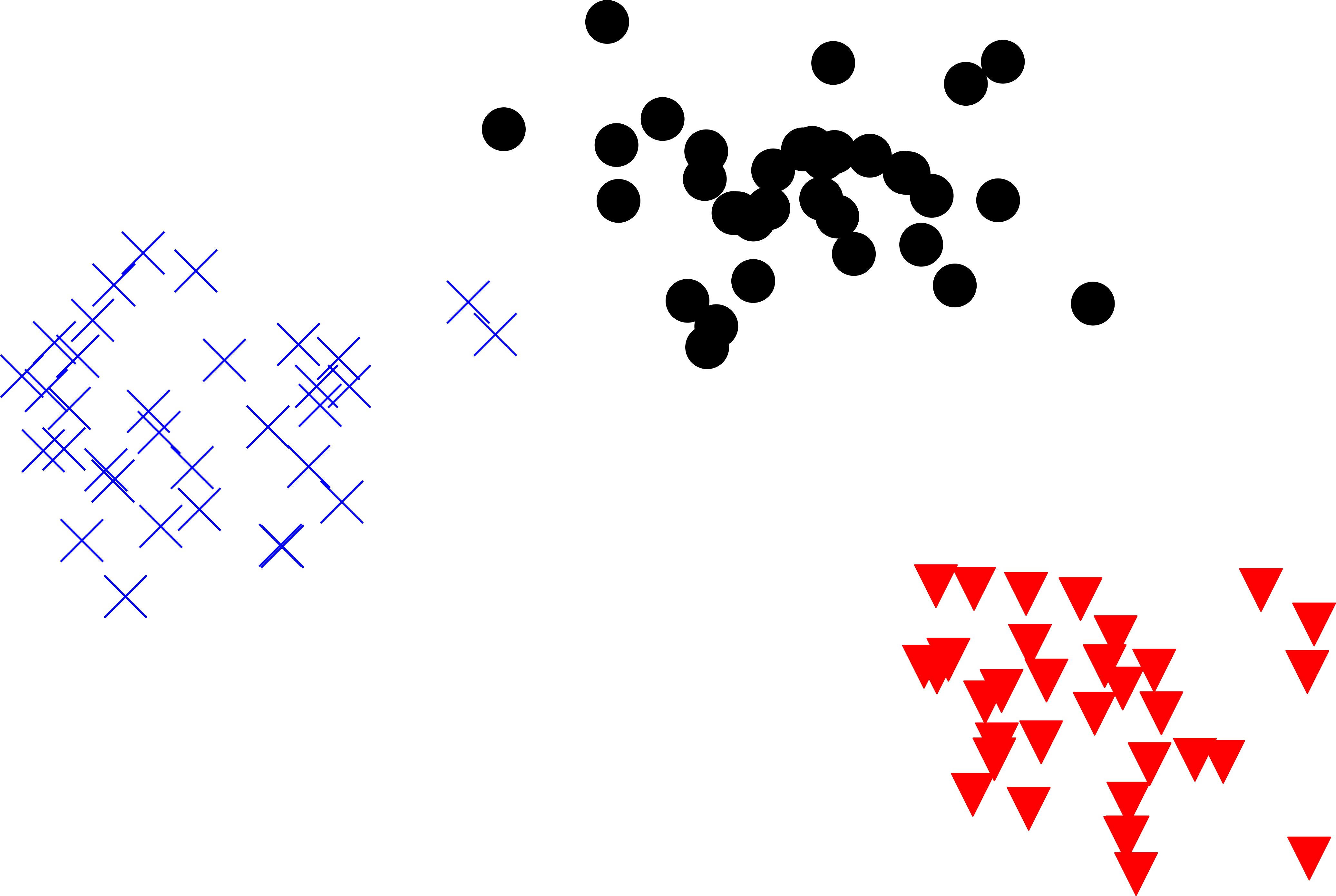}
        \label{sfig:nonparametric_gemini}    
    }
    \end{center}
    \subfloat[Snippet of code to reproduce the example]{
        % subfloat does not like verbatim, need to pass through external file: https://tex.stackexchange.com/questions/34946/how-can-i-put-lstlisting-block-into-subfloat-block
\begin{lstlisting}[language=PythonPlus, style=colorEX]
from sklearn import datasets
X, _ = datasets.make_blobs(centers=3, cluster_std=0.5, random_state=0)
# Take a nonparametric model and fit it using MI as objective
from gemclus.nonparametric import CategoricalModel
mi_model = CategoricalModel(n_clusters=3, gemini="mi")
y_pred = mi_model.fit_predict(X)
# Same nonparametric model, optimise a distance-regularised MI
gemini_model = CategoricalModel(gemini="mmd_ova", random_state=0,
    learning_rate=1e-2)
y_pred_regularised = gemini_model.fit_predict(X)
\end{lstlisting}
        \label{sfig:listing_nonparametric}
    }
    \caption{Clustering of mixture of 3 isotropic Gaussian distributions by a nonparametric model. When the cluster memberships are directly optimised by mutual information, the clusters do not correspond to any of the initial Gaussian distributions because the cluster memberships are not derived from the position of the samples. In constrast, MI regularised by maximum mean discrepancy can optimise correctly the nonparametric model. This example is taken from \citet{ohl_generalised_2022} and can be reproduced using their package GemClus~\citep{ohl_gemclus_2023}.}
    \label{fig:nonparametric_gaussian_clustering}
\end{figure}

However, multiple works in the past and still today point to the inexpressivity of local maxima of MI~\citep{bridle_unsupervised_1992, corduneanu_information_2002}. In essence, inexpressivity means here a maximised mutual information with respect to the parameters of the model does not translate into good clustering. For example, \citet{ver_steeg_demystifying_2014} showed with a mixture of uniform distributions how MI fails to properly separate the initial bins, and \citet{corduneanu_information_2002} empirically showed that the associated clusters to samples do not affect MI as long as the clusters are balanced. We can also give the example of \citet{ohl_generalised_2022} in Figure~\ref{fig:nonparametric_gaussian_clustering} where a nonparametric model is trained using mutual information. This nonparametric model takes the form:
\begin{equation}\label{eq:nonparametric_model}
    \p(y=k\mid\x=\x_i) = \tau_{ki}.
\end{equation}

In other words, a cluster membership for the cluster $k$ to the $i$-th sample is directly assigned. The ``nonparametric'' model has $N\times K$ parameters: $\theta = \{\tau_{ki}\}_{i=1, k=1}^{n,K}$ that are constrained such that $\sum_{k=1}^K \tau_{ki}=1$ and $\tau_{ik}$ is positive. These cluster memberships $\tau_{ik}$ do not depend on the value of the sample $\x_i$ and only match $\x_i$ by indexing. Consequently, the nonparametric model creates only a partition of the dataset, but it cannot generalise to unseen samples.

Due to the absence of relationship to the values of the samples in the model parameters, their optimisation with regard to mutual information produces clusters that plainly create a balanced partition of the dataset, regardless of its shape \cite[Corollary 1]{ohl_generalised_2023}. This limitation of MI justifies the usage of regularisations to constrain sufficiently the clustering models. We will see that these regularisations were part of the discriminative clustering framework from the very start in Section~\ref{ssec:thrive_beginning}. To the best of our knowledge, few works did question the quality of MI as an objective in itself. Among them, \citet{ohl_generalised_2022} proposed to replace the Kullback-Leibler divergence $\kl$ by any other distance $D$:
\begin{equation}\label{eq:gemini}
    \gemini[D]{\x;y} = \expectation{y \sim \p(y)}{D\left(\p(\x\mid y) \| \pdata(\x)\right)},
\end{equation}

\noindent preferably using geometry-based distances like the Wasserstein distance ~\citep{peyre_computational_2019} or the maximum mean discrepancy~\citep{gretton_kernel_2012}. An example of non-geometry-based distance is the $\chi^2$ divergence, which was explored by \citet{sugiyama_information_2011}. This objective, named \emph{generalised mutual information} (GEMINI), tackles the inexpressivity of local maxima by enforcing the hypothesis that samples that are close to each other should remain in the same cluster. The GEMINI objective preserves the key property of independence and null value of the initial MI.

\section{Other clustering models}
\label{sec:other}

Between generative and discriminative models, there exist other clustering models. We discuss some of these objectives and compare their (dis)advantages with mutual information.

\subsection{K-means}
\label{ssec:other_kmeans}

K-means is the most standard clustering algorithm~\citep{likas_global_2003}. The algorithm consists of two alternating steps. Starting from a set of $K$ centroids: $\{\vec{\mu}_k\}_{k=1}^K$, all samples in the dataset are assigned to the cluster $\cluster_k$ matching their closest centroid $\vec{\mu}_k$. Then each centroid is recomputed as the mean of all samples assigned to the corresponding cluster. Overall, the K-means algorithm minimises the following objective~\citep{lloyd_least_1982, elkan_using_2003}, although there are no guarantees of reaching the global optimum:
\begin{equation}
    \vec{\mu}_{1\ldots K}^\star \in \argmin_{\vec{\mu}_{1\ldots K}}\sum_{k=1}^K \sum_{\x \in \cluster_k}\norm{\x - \vec{\mu}_k}_2^2.
\end{equation}

The simplicity of K-means makes it a standard basis for clustering algorithms.
However, its decision boundaries are linear because they appear for all samples that lay equidistantly between two close means. Consequently, this does not allow complex cluster shapes when the Euclidean distance is used between data samples. An alternative called kernel K-means~\citep{dhillon_kernel_2004} can be used to alleviate more complex boundaries. Instead of drawing a boundary in the Euclidean space, it is drawn in a reproducing kernel Hilbert space $\rkhs$ endowed with projection $\varphi$~\citep{hofmann_kernel_2008}. The new objective is described:
\begin{equation}\label{eq:generative_kmeans}
    \objective(\{\vec{\mu}_k\}_{k=1}^K) = \sum_{k=1}^K \sum_{\x \in \cluster_k} \norm{\varphi(\x) - \vec{\mu}_k}_\rkhs^2.
\end{equation}

Beyond its wide applicability, K-means offers multiple advantages. First, the definition of fixed centroids also allows the clustering of unseen samples, \ie samples that were not part of the training dataset. Second of all, this algorithm is fast as it scales linearly to the number of samples $n$ and the number of input features $d$. However, K-means has several drawbacks: it is very sensitive to outliers, \ie samples far away from the main modes of the data, and it displays high sensitivity to the initialisation. Moreover, the clustering model conveyed by K-means is a mixture of Dirac models, which does not provide a fine-grained quantification of the cluster conditional probability. This is often referred to as \emph{hard clustering}.

If we disregard the hard membership assignments, K-means can be seen as a generative clustering model as it is similar to a mixture of isotropic Gaussian distributions. Indeed, both algorithms produce linear decision boundaries. The K-means score can also be seen and used as an objective function for discriminative clustering, where the model is non-parametric as in Eq.\ref{eq:nonparametric_model}. Indeed, \citet{ohl_kernel_2024} showed that by using the equality:
\begin{equation}
    \sum_{\x \in \cluster_k} \norm{\varphi(\x) - \underbrace{\inv{\card{\cluster_k}}\sum_{\vec{y} \in \cluster_k} \varphi(\vec{y})}_{\vec{\mu}_k}}_\rkhs^2 = \inv{2\card{\cluster_k}} \sum_{\x, \vec{y} \in \cluster_k} \norm{\varphi(\x) - \varphi(\vec{y})}_\rkhs^2.
\end{equation}

\noindent and the kernel trick with a kernel function $\kappa$, the kernel K-means objective to minimise becomes, constants aside:\begin{equation}\label{eq:discriminative_kmeans}
    \objective(\{\cluster_k\}_{k=1}^K) = - \sum_{k=1} \frac{\sum_{\x, \vec{y} \in \cluster_k} \kappa(\x, \vec{y})}{\card{\cluster_k}}.
\end{equation}

This objective is now a function of the partition, \ie the clustering, instead of parametric centroids $\vec{\mu}_k$ that are iteratively updated. An equivalent formulation using can be found in \citet{franca_kernel_2020}.

\begin{remark} The Equation~(\ref{eq:generative_kmeans}) is often referred to as the average \emph{intra-cluster similarity} although it uses distances~\citep{soler_data_2013}, the objective from Eq.~(\ref{eq:discriminative_kmeans}) could be described as the average cluster kernel, which is different from distances or similarities.
\end{remark}

\subsection{Spectral clustering}
\label{ssec:other_spectral}

Related to kernel K-means, spectral clustering is a multi-step algorithm~\citep{von_luxburg_tutorial_2007} in which an alternative representation of the data is derived and used as input for a K-means algorithm. At the start of the algorithm, an affinity matrix is computed between all samples. The choice of the notion of affinity is arbitrary: $\epsilon$-neighbourhood, $n$ nearest neighbours, or a kernel like radial basis function (RBF), polynomial. This affinity matrix allows us to define a Laplacian matrix that views the complete dataset as a graph. The nature of the Laplacian can vary, \eg being normalised or symmetrised~\citep{shi_normalized_2000, ng_spectral_2001}. Then, an alternative representation of the data emerges from the spectrum of this Laplacian matrix. Specifically, the first $K$ eigenvectors of the Laplacian matrix are used as new features. Consequently, each sample is associated with a vector of dimension $K$ where its $k$-th component is its weight for the $k$-th eigenvector.

While spectral clustering is efficient when cluster structures are non-convex, its strong requirement is the construction of a graph between all samples which enforces the fusion of modalities when several modalities are present. Moreover, the identification of the spectrum of the Laplacian prevents the model from generalising to unseen samples. Once again, due to the final K-means algorithm, the spectral clustering can be seen as a delta Dirac distribution which only delivers hard membership to clusters for all samples.

\section{Thrive of mutual information: from clustering to representation learning}
\label{sec:thrive}

Mutual information is an elegant objective function that can train through gradient ascent various differentiable models as we showed in Section~\ref{ssec:challenge_mutual_information}. We trace here the multiple usages it has met for the past 30 years in the context of clustering.

\subsection{Early usage of mutual information for clustering}
\label{ssec:thrive_beginning}

To the best of our knowledge, MI was first used as an objective for learning discriminative clustering models by \citet{bridle_unsupervised_1992}. They described MI as an objective that maximises the \emph{fairness} of a model, \ie the entropy of the cluster proportions, and aim to maximise \emph{firmness}, \ie minimising the conditional entropy. By rewriting mutual information, we have:
\begin{equation}
    \mi{\x;y} = \underbrace{\entropy{y \mid \x}}_\text{Firmness} - \underbrace{\entropy{y}}_\text{Fairness}.
\end{equation}

Thus, a good clustering model is fair but firm.

Two decades later, \citet{krause_discriminative_2010} initiated again the work on discriminative clustering models with mutual information. Similarly to \citet{bridle_unsupervised_1992}, they first propose to train a logistic regression but specifically add a $\ell_2$ constraint  to alleviate the constraint on the firmness of the classifier model. Indeed, in line with the criticisms from Section~\ref{ssec:challenge_mutual_information}, a model that is too firm, \ie has very low conditional entropy, produces overconfident clustering probabilities, and so should the $\ell_2$ regularisation avoid an excessive firmness. Among the multiple proposals of this framework called regularised mutual information (RIM), \citet{krause_discriminative_2010} also propose to learn a regularised logistic model on a positive semi-definite kernel matrix of some dataset instead of the sample features to get a non-linear decision boundary.

These two approaches decouple the model $\p(y\mid \x)$ that makes assumptions about the shape of the decision boundary from the objective. Such methods are proposed and implemented in the GemClus software~\citep{ohl_gemclus_2023}.

Another interesting approach, named the information bottleneck, focuses on the notion of information relevance. Indeed: \textit{``The problem of extracting a relevant summary of data, a compressed
description that captures only the relevant or meaningful information, is not
well-posed without a suitable definition of relevance'' \citep{tishby_information_2000}.}

The information bottleneck framework is derived from signal processing theory and is related to rate-distortion theory~\citep{tishby_information_2000}. As such, it considers an input variable $Z$, that must be quantised into a lower-dimensional or discrete variable called code $\tilde{Z}$. However, this quantisation must be done such that another output variable $\bar{Z}$, can be recovered from the code. The final model is written:
\begin{equation}
    p^\star(\tilde{Z}\mid Z) \in \argmin_{p(\tilde{Z}\mid Z)} \mi{\tilde{Z};Z} - \beta \mi{\tilde{Z};\bar{Z}}.
\end{equation}

The hyperparameter $\beta$ controls the trade-off between a highly detailed compression ($\beta=\infty$) and an independent compression ($\beta=0$) of $Z$ into $\tilde{Z}$~\citep{tishby_information_2000}. 
The distribution we seek to optimise $p(\tilde{Z} \mid Z)$ can be parametric or nonparametric.

Initially, the information bottleneck was designed in a supervised context: \citet{tishby_information_2000} explicitly assumes that they have access to some distribution $p(Z, \bar{Z})$. In this context, we can interpret the input variable $Z$ as the data $\x$ and the output variable $\bar{Z}$ as the targets, \ie observed labels, $l$. Only the quantised variable $\tilde{Z}$ remains to be defined. We can define this variable as the cluster membership that we seek to enhance classification. In other words: we seek to compress the data into several clusters that we then use to alleviate the classification task. The corresponding objective is:
\begin{equation}
\objective = \mi{\x;y} - \beta \mi{l;y},
\end{equation}

For example, \citet{slonim_document_2000} used this objective to discover word clusters using documents as inputs $\x$, words as classes $l$ and word clusters $y$ to compress the documents. Similarly, \citet{dhillon_divisive_2003} proposed using word clusters to improve document classification. In their case, the targets $l$ are labels that should be assigned to documents instead of words. However, they framed the information bottleneck differently. If we seek more clusters than classes, the targets can then seem as more compressed representation than the clusters. \citet{dhillon_divisive_2003} permute the role of classes and clusters in the information bottleneck and optimise:

\begin{equation}
    \objective = \mi{l; \x} - \mi{l; y}.
\end{equation}

Thus, the information bottleneck compresses the words into clusters $y$ that suffice to guess the class of documents. Notice that in this context, the hyperparameter $\beta$ was set to 1. \citet{slonim_information-based_2005} adapted as well this framework for doing only clustering and without the goal of recovering a target contained in the dataset. To that end, they replaced the first mutual information by the expectation of the cluster similarity measure \texttt{sim}. Taking a random subset of $r$ samples per cluster to evaluate the similarity, their objective is:
\begin{equation}
    \objective = \expectation{y\sim\p(y)}{\expectation{\x_{1\cdots r} \sim \prod_{i=1}^r \p(\x_i\mid y)}{\texttt{sim}(\x_{1\cdots r})}} - \beta \mi{\x;y}.
\end{equation}

It is interesting to see that in this specific example, one of the mutual information objectives was replaced to take into account distances between samples of clusters to regulate mutual information between the clusters and the data. This joins the observations of \citet{ohl_generalised_2022} and their introduction of distances to regularise mutual information itself. This introduction of a notion of similarity will play a key role in entangling clustering hypotheses in mutual information, as we will see in Section~\ref{ssec:thrive_contrastive}. In contrast to the work of \citet{bridle_unsupervised_1992} and \citet{krause_discriminative_2010}, the early works using the information bottleneck framework used nonparametric models~\citep{dhillon_divisive_2003, slonim_information-based_2005}, which we defined in Eq.~(\ref{eq:nonparametric_model}).

\subsection{Towards deeper networks}
\label{ssec:thrive_deeper}

Two years after the RIM model, \citet{krizhevsky_imagenet_2012} introduced the AlexNet model, an example of strong modern deep learning success. Inspired from this success, depth also affected the neural networks involved in clustering tasks. For instance, \citet{xie_unsupervised_2016} proposed to use autoencoder architectures combined with K-means clustering in the compressed space to benefit from nonlinear architectures. To the best of our knowledge, the first deep clustering model involving mutual information was proposed by \citet{hu_learning_2017} who replaced the $\ell_2$ constraint of \citet{krause_discriminative_2010} by a virtual adversarial constraint~\citep{miyato_virtual_2018}. The key idea is to add small perturbations to the input sample and ensure that the assigned cluster is consistent with the clustering of the initial sample:
\begin{equation}
\theta^* \in \argmax_\theta \mi{\x;y} \quad \text{s.t.} \quad \kl[\p(y\mid \x)][\p(y\mid\x + \vec{\varepsilon})]\approx 0,
\end{equation}

\noindent where $\vec{\varepsilon}$ is a very small noise.
Augmentations such as random cropping, scaling, or rotations were also used to perturb the initial sample. The key idea is that these perturbations provide invariances in clustering~\citep{ji_invariant_2019}. This can also be seen as designing a specific neighbourhood in which we can sample the perturbation $\vec{\varepsilon}$.

Deep clustering methods progressively used deeper and deeper networks, \eg ResNets~\citep{he_deep_2016}. Moreover, they sought to improve architectures by providing regularisations through auxiliary loss terms~\citep{ren_deep_2024}, \eg auxiliary clustering heads~\citep{ji_invariant_2019}. As an example of regularisation impact: maximising MI with $\ell_2$ constraint can be equivalent to a soft and regularised K-Means in a feature space~\citep{jabi_deep_2019}. This drift towards deep networks and the introduction of data augmentation led to the emergence of contrastive learning as a core basis for deep discriminative clustering methods. We now detail its relationship with mutual information.

\subsection{From discrete to continuous output variables: contrastive learning and infoMax}
\label{ssec:thrive_contrastive}

\subsubsection{Mutual information lower bounds}

With the introduction of deep neural networks in clustering, the field of \emph{deep clustering} rapidly gave way to the development of discriminative models that predict a continuous variable $\z \in \mathcal{Z}$ instead of a simple cluster assignment $y \in \naturalto{K}$. These models are no longer designed for clustering, but are compatible with clustering in the space of learnt representations. For example, a simple continuous distribution based on a discriminative network could be:
\begin{equation}\label{eq:example_representation_model}
    \p(\z\mid\x) = \mathcal{N}(\z\mid\psi_\theta(\x), \sigma^2\pmb{I}).
\end{equation}

However, we cannot optimise representation learning model parameters directly using mutual information. In Section~\ref{sec:challenge}, we benefited from the discrete nature of the target $y$ that yielded a finite sum to estimate mutual information. Now, our target variable is a continuous representation $\z$, which implies that we must compute an intractable integral to get $\p(\z)$. As a workaround, lower bounds of the mutual information between $\x$ and $\z$ are used as objective functions to learn such models.

Several lower bounds have been designed, as named by \citet{poole_variational_2019}: MINE~\citep{belghazi_mutual_2018}, NCE~\citep{van_den_oord_representation_2018}, BA~\citep{barber_im_2003}. We offer some insight on their derivation in the next paragraphs. For more in-depth details regarding the derivations, we refer to \citet{poole_variational_2019}.

To get a lower bound, let us express mutual information using $p(\x\mid \z)$. Since we do not know its value by our discriminative assumptions, we will try to approximate this distribution using a proposal distribution $q(\x\mid \z)$. This allows us to rewrite MI~\citep{barber_im_2003} as:
\begin{align}
    \mi{\x;\z} &= \expectation{p(\z)}{\kl[p(\x\mid \z) \frac{q(\x \mid \z)}{q(\x \mid \z)}][p(\x)]},\\
    &= \expectation{p(\x, \z)}{\log q(\x\mid \z)} +\entropy{\x} +\expectation{p(\z)}{\kl[p(\x\mid \z)][q(\x\mid \z)]},\\
    &\geq \expectation{p(\x, \z)}{\log q(\x\mid \z)} +\entropy{\x}.\label{eq:initial_lower_bound}
\end{align}

In order to connect this expression to other lower bounds, \citet{poole_variational_2019} suggest to take $q$ from the energy-based variational family:
\begin{equation}
q(\x\mid\z) = \frac{\pdata(\x)}{Z(\z)} e^{E(\x,\z)},
\end{equation}

\noindent where $Z(\z)=\expectation{\pdata(\x)}{e^{E(\x,\z)}}$.

Integrating this proposal distribution in our lower bound cancels the entropy term on $\x$, and we obtain:
\begin{equation}\label{eq:basic_lower_bound}
    \mi{\x;\z} \geq \expectation{p(\x, \z)}{\log \frac{e^{E(\x,\z)}}{Z(\z)}}
\end{equation}

Evaluating this lower bound requires the value of the denominator $Z(\z)$. The suggestion of \citet{poole_variational_2019} is to directly use Monte Carlo:
\begin{equation}\label{eq:monte_carlo_constant}
    Z(\z) \approx \inv{n} \sum_{i=1}^n e^{E(x_i, \z)}.
\end{equation}

Using such an estimate requires that we have access to a batch of $n$ i.i.d. samples $\x_i$. This has been our main assumption so far in discriminative clustering. We can here benefit from the independence assumption to further enhance our estimator of the mutual information lower bound. Indeed, mutual information is invariant to the addition of a variable $\vec{v}$ independent of $\x$ and $\z$, noted: $\mi{\x,\vec{v};\z} = \mi{\x;\z}$. Consequently, looking at each sample as individual independent random variables and detailing the probability distribution over all samples yields the following:
\begin{equation}\label{eq:separation_mi}
    \mi{\x_1; \z} = \mi{\x_1, \x_2, \ldots, \x_n; \z}.
\end{equation}

We now assume that we have i.i.d. pairs of variables $(\x_i, \z_i)$. Owing to this assumption, we can estimate mutual information by summing the mutual informations between each representation $\z_i$ and each individual samples $\x_i$. Each term of this sum has exactly the same value:
\begin{equation}\label{eq:multisample_mi}
    \mi{\x_1, \ldots, \x_n; \z} = \inv{n} \sum_{i=1}^n \mi{\x_i; \z_i}.
\end{equation}

We can wrap up the mutual information in an expectation over the distribution of all remaining samples. Thus:
\begin{equation}
    \mi{\x_1; \z} = \inv{n} \sum_{i=1}^n \expectation{\Pi_{j\neq i}^n \pdata(\x_j)}{\mi{\x_i; \z_i}}.
\end{equation}

Each expectation can be seen as the tool to get our Monte Carlo samples for estimating $Z(\z)$ from Eq.~(\ref{eq:monte_carlo_constant}). We then incorporate the lower bounds from Eq.~(\ref{eq:basic_lower_bound}) for each mutual information term:
\begin{equation}\label{eq:info_nce}
    \mi{\x;\z} \geq \inv{n}\sum_{i=1}^n\expectation{\pdata(\x_1,\ldots, \x_n) \p(\z\mid \x_i)}{ \log  \frac{e^{E(\x_i,\z_i)}}{\sum_{j=1}^n e^{E(\x_j,\z_i)}}} + \log{n}.
\end{equation}

The proof that this is an actual lower bound can be found in \citet{poole_variational_2019}. Note that this demonstration is also valid for a discrete variable $y$ instead of $\z$.

Notice that the constant $\log{n}$, which comes from our Monte Carlo estimate of the denominator, introduces a bias in the lower bound. In this final expression, the samples $\x_{1,\ldots,n}$ that we use to estimate our initial denominator $Z(\z)$ are the same ones from the outer expectation. This lower bound as the Info NCE and can be linked to the temperature-scaled cross-entropy of contrastive learning (NT-XENT, \citealp{chen_simple_2020}).

There exist an optimal distribution for our variational proposal. It is the energy as the log of the output of the model:
\begin{equation}
    E(\x,\z) = \log p(\z \mid \x).
\end{equation}

Consequently, the proposal distribution $q(\x\mid \z)$ becomes equal to $\p(\x \mid\z)$. If we insert this proposal directly into the lower bound from Eq~(\ref{eq:info_nce}), we finally have an objective for training our parameters $\theta$ with a continuous output for the model:
\begin{equation}\label{eq:exact_bound}
    \mi{\x;\z} \geq \inv{n}\sum_{i=1}^n\expectation{\pdata(\x_1,\ldots, \x_n) \p(\z\mid \x_i)}{ \log  \frac{\p(\z_i\mid \x_i)}{\sum_{j=1}^n \p(\z_i\mid \x_j)}} + \log{n}.
\end{equation}

In this equation, our estimate of the denominator $Z(\z)$ directly corresponds to the estimate of $p(\z)$. Although, Eq.~(\ref{eq:exact_bound}) seems like a compelling objective to train representation models, \ie $\p(\z\mid\x)$, it would not be reasonable to directly use it because the estimate of $p(\z)$ is precisely the limitation we sought to avoid. We initially motivated at the beginning of this section that the continuous, and potentially high-dimension nature of $\z$ is responsible of the intractability of $p(\z)$.

Now that we have a lower bound on MI that we can potentially use in high-dimensional spaces, we need to adapt our model for clustering instead of predicting a continuous representation. An easy solution would be to use a second clustering algorithm to cluster the samples in the space $\mathcal{Z}$. Although this is a valid approach, this implies our model $p_\theta(\z\mid \x)$ is not any longer the one to achieve clustering. To maintain the clustering nature of our model, we need to consider differently the meaning of the continuous representation $\z$. Instead of considering $\z$ as an output from the model we want to learn, we will use it as a joint property of $\x$ using augmentations: we turn to contrastive learning.

\subsubsection{Contrastive learning}\label{sssec:contrastive_learning}

Contrastive learning is an integral part of representation learning through the lens of self-supervised learning. Representation learning consists in finding high-level features $\z_i$ extracted from the data $\x_i$ to perform a \emph{downstream task}, \eg clustering or classification. However, the nature of the model is different with regards to mutual information in Eq.~(\ref{eq:info_nce}).

The key idea of contrastive learning is to perform a set of random augmentations on a sample $\x_i$, then maximise the similarity between the representation associated with this sample and its augmentation, while decreasing the similarity with any other sample. This choice implies that we \emph{no longer} maximise the mutual information between the data $\x$ and the representation variable $\z$, but a variable corresponding to the augmentation of the data: $\text{Aug}(\x)$ \cite{do_clustering_2021}. This also means that the conditional distribution $p(\text{Aug(\x)}\mid \x)$ is unknown. Nevertheless, we assume that we can sample augmentations of $\x$ easily. We can rewrite Eq.~(\ref{eq:info_nce}) in the context of contrastive clustering as:
\begin{equation}\label{eq:ntxent}
    \mi{\x; \text{Aug}(\x)} \geq \inv{n}\sum_{i=1}^n\expectation{p(\x_1,\ldots, \x_n) p(\text{Aug}(\x)\mid \x_i)}{\log \frac{e^{E_\text{constrastive}(\x_i,\text{Aug}(\x_i))}}{\sum_{j=1}^n e^{E_\text{constrastive}(\x_i,\text{Aug}(\x_j))}}} + \log{n}.
\end{equation}

In this context, the distributions on which we perform the expectations are not parameterised \emph{any longer} by $\theta$. In fact, mutual information between $\x$ and $\text{Aug}(\x)$ is constant because we fixed an augmentation strategy. The parameters of our neural networks are now \emph{hidden in the critic} function $E_\text{contrastive}$, which must favour high similarities between samples and their respective augmentations:
\begin{equation}
    E_\text{constrastive}(\x_i,\text{Aug}(\x_j)) = \texttt{sim}\left(\psi_\theta(\x_i), \psi_\theta(\text{Aug}(\x_j))\right).
\end{equation}

Our model does not aim any longer at maximising mutual information $ \mi{\x; \text{Aug}(\x)}$, which is a constant, but try to approximate a distribution. Indeed, in our initial lower bound from Eq.~(\ref{eq:initial_lower_bound}), the optimal solution is to get $q(\x\mid \z)$ equal to $p(\x\mid \x)$, in which case the first term becomes the entropy of $\x$ given $\z$ and the sum recovers the exact MI. That is why contrastive learning is about learning the distribution $p(\x\mid \text{aug}(\x))$ using the approximate $q(\x\mid \text{aug}(\x))$.

Contrastive learning provides a convenient framework that takes advantage of the key idea that the neighbourhood of a sample should remain in the neighbourhood of the representation of this sample. The nature of the neighbourhood is dependent on the choices of data augmentation. Although it is not related to clustering, this approach indirectly views each sample in an individual cluster because we maximise MI between two variations of $\x$. This means that the only elements that can go in the same cluster are the augmentations of the samples and are called \emph{positive pairs}, while all the others, including their augmentations, are called \emph{negative pairs} and must be excluded.
However, such setup completely loses the end goal of clustering that is to put samples into $K$ categories. Therefore, additional tricks are required to bring back the continuous representations to a clustering model. We give now some examples of such tricks. Note that we purposefully omit some details for the sake of clarity, especially regarding how augmentations are handled through batches during learning and regularisations.

In the spirit of the SIMCLR model~\citep{chen_simple_2020}, \citet{do_clustering_2021} propose to decompose the model in two different parts. One is a backbone $\psi$ learning common representations for the second part that comprises two different projection heads: one for clustering $\chi$, \ie a softmax-ended function, and $\varphi$ a projection to the continuous domain. Note that we omit the parameters for brevity. The model thus comprises two functions, the clustering function:
\begin{equation}
    \begin{array}{r l}
        \chi \circ \psi : \dataspace &\mapsto \Delta^{K-1}, \\
         \x_i &\mapsto \chi \circ \psi (\x_i)=y_i.
    \end{array}
\end{equation}

\noindent and the representation function:
\begin{equation}
    \begin{array}{r l}
        \varphi \circ \psi : \dataspace &\mapsto \mathcal{Z}, \\
         \x_i &\mapsto \varphi \circ \psi (\x_i)=\z_i.
    \end{array}
\end{equation}

Then, by summing two mutual informations with different critics $E_1$ and $E_2$, one for the representation and one for the clustering, \citet{do_clustering_2021} achieve a model with features presenting high intra-cluster variability and low inter-group similarity:
\begin{equation}
    \objective = \underbrace{\gemini[E_1]{\x; \text{Aug(\x)}}}_{E_1(\cdot, \cdot) = \texttt{sim}(\varphi\circ\psi(\cdot), \varphi\circ\psi(\cdot))} + \underbrace{\gemini[E_2]{\x; \text{Aug(\x)}}}_{E_2(\cdot, \cdot) = \texttt{sim}(\chi\circ\psi(\cdot), \chi\circ\psi(\cdot))}.
\end{equation}

% \begin{wrapfigure}{r}{0.5\linewidth}
%     \centering
%     \includegraphics[width=\linewidth]{imgs/deep_notations.pdf}
%     \caption{Graphical explanation of notations used for deep clustering models.}
%     \label{fig:deep_notations}
%     \vspace{-\baselineskip}
% \end{wrapfigure}
\begin{figure}[t]
    \centering
    \includegraphics[width=0.7\linewidth]{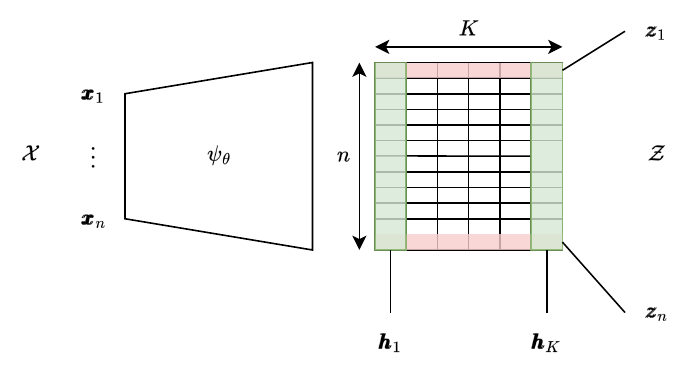}
    \caption{Graphical explanation of notations used for deep clustering models.}
    \label{fig:deep_notations}
    %\vspace{-\baselineskip}
\end{figure}

Another line of work proposes instead to construct a critic function $E$ that evaluates the similarity between the distribution of the clusters rather than the representations. We describe the notations in Figure~\ref{fig:deep_notations}. Let us note the conditional distribution as a vector:
\begin{equation}
    \vec{h}_k^\top = [\p(y=k\mid \x_1), \ldots, \p(y=k\mid \x_i), \ldots, \p(y=k\mid \x_n)] \in \real^n.
\end{equation}

Using this formulation, \citet{huang_deep_2020} proposed to maximise $\mi{\vec{h}_{1, \ldots K}; \bar{\vec{h}}}$ where $\bar{\vec{h}}$ is the cluster distribution obtained after applying random augmentations on the samples. \citet{li_contrastive_2021} extended this idea by adding a second mutual information between the representations, as we explained for \citet{do_clustering_2021}.

The example of \citet{huang_deep_2020} offers a different perspective on the usage of MI. Instead of constraining the neural network output with a softmax to obtain the probabilities of a discrete variable $y$, we can directly seek to optimise MI between samples and a set of $K$ cluster centroids $\vec{\mu}_k$. We assume that the centroids are i.i.d. and can therefore chain two times in a row the multi-sample definition of MI from Eq.~(\ref{eq:multisample_mi}):
\begin{equation}
    \mi{\x_{1,\ldots,n}; \vec{\mu}_{1\ldots, K}} = \inv{K}\sum_{k=1}^K \expectation{\Pi_{l\neq k}^K p(\vec{\mu}_l)}{\inv{n} \sum_{i=1}^n \expectation{\Pi_{j\neq i}^n p(\x_j)}{\mi{\x_i;\vec{\mu}_k}}}.
\end{equation}

In practice, the centroids $\mu_k$ are sampled them Dirac distributions whose locations are learnt. We then have the freedom of choosing either the proposal distribution $q(\vec{\mu}\mid \x)$, in which case the normalisation constant must be estimated using the centroids $\mu_k$, or the proposal $q(\x \mid \vec{\mu})$, in which case we estimate the normalisation constant using the batch of samples $\x_i$. Let us choose the former. We obtain a new lower bound on MI that seeks to connect a batch of samples to a batch of centroids:
\begin{equation}
    \mi{\x_{1,\ldots,n}; \vec{\mu}_{1\ldots, K}} \geq \inv{K}\sum_{k=1}^K \expectation{\Pi_{l\neq k}^K p(\vec{\mu}_l)}{\inv{n} \sum_{i=1}^n \expectation{p(\vec{\mu}_k\mid \x_i) p(\x_1, \ldots, \x_n)}{\log \frac{e^{E(\x_i,\vec{\mu}_k)}}{\sum_{l=1}^K e^{E(\x_i,\vec{\mu}_l)}}}}.
\end{equation}

This new lower bound on MI corresponds to the objective of representation models using swapping assignments~\citep{caron_unsupervised_2020}. To show this link, let us consider now that our samples $\x_i$ are in fact $n$ different augmentations of the same original sample, noted $\x_i \mid \tilde{\x} \equiv \text{Aug}^{(i)}(\tilde{\x})$. This addition holds with our statistical assumptions because the samples $\x_i$ remain i.i.d. given  $\tilde{\x}$. We can then rewrite the lower bound on MI by adding the condition on $\tilde{\x}$ and get:
\begin{equation}
    \mi{\x_{1,\ldots,n} \mid \tilde{\x}; \vec{\mu}_{1\ldots, K}} \geq \inv{K}\sum_{k=1}^K \expectation{\Pi_{l\neq k}^K p(\vec{\mu}_l)}{\inv{n} \sum_{i=1}^n \expectation{p(\vec{\mu}_k\mid \x_i, \tilde{x} ) p(\x_1, \ldots, \x_n\mid \tilde{\x})}{\log \frac{e^{E(\x_i,\vec{\mu}_k)}}{\sum_{l=1}^K e^{E(\x_i,\vec{\mu}_l)}}}}.
\end{equation}

Owing to this additional condition on $\tilde{x}$ and the independence of all samples $\x_j$ to $\x_i$ and $\mu_k$, we can use the product rule of probabilities to change the distributions on which the innermost expectation is done. Thus:
\begin{align}
    p(\vec{\mu}_k\mid \x_i, \tilde{\x} ) p(\x_1, \ldots, \x_n\mid \tilde{\x}) &= p(\vec{\mu}_k, \x_1, \ldots, \x_n \mid \tilde{\x})\\
    &= p(\vec{\mu}_k, \x_1, \ldots, \x_{i-1}, \x_{i+1}, \ldots, \x_n \mid \tilde{\x}, \x_i)p(\x_i \mid \tilde{\x}),\\
\end{align}

Finally, this rewritten distribution gives us a generalised swapping representation model for $n$ augmentations that we seek to map to $K$ centroids:
\begin{equation}
    \mi{\x_{1,\ldots,n} \mid \tilde{\x}; \vec{\mu}_{1\ldots, K}} \geq \inv{K}\sum_{k=1}^K \expectation{\Pi_{l\neq k}^K p(\vec{\mu}_l)}{\inv{n} \sum_{i=1}^n \expectation{p(\x_i\mid \tilde{\x})p(\vec{\mu}_k, \x_1, \ldots, \x_{i-1}, \x_{i+1}, \ldots, \x_n \mid \tilde{\x}, \x_i)}{\log \frac{e^{E(\x_i,\vec{\mu}_k)}}{\sum_{l=1}^K e^{E(\x_i,\vec{\mu}_l)}}}}.
\end{equation}

The example of \citet{caron_unsupervised_2020} corresponds to only two augmentations ($n=2$). In their work, the distribution $p(\vec{\mu}_k, \x_j \mid \tilde{x}, \x_i)$, $j\neq i \in \{1,2\}$ is estimated with a regularised optimal transport problem on a batch of $B$ samples $\tilde{\x}$, and the energy function $E$ is the temperature-scaled similarity between the centroid $\vec{\mu}_k$ and the representation of the augmented sample by the neural network $\psi_\theta(\x_i)$.

To conclude, the advantage of contrastive learning is the single-stage nature of training. However, the performances of these methods are tied to the choices of augmentations, which may not always be clear cut depending on the data. For example, basic transformations such as translations, Gaussian blur, and scaling were shown to be efficient augmentations for classification and segmentation tasks of CT scans~\citep{chlap_review_2021, garcea_data_2023}. Yet, this does not imply that these augmentations could be beneficial in an unsupervised context.

\subsubsection{The InfoMax principle}

Aside contrastive clustering, the InfoMax principle is another framework involving mutual information for learning continuous representations. It was started by \citet{linsker_self-organization_1988} and its goal was to construct a network such that: \textit{``The information that reaches a layer is processed so that the maximum amount of information is preserved. We have seen that this does not in general lead to a trivial one-to-one identity mapping [...]'' \citep{linsker_self-organization_1988}}.

This principle was later refined by \citet{hjelm_learning_2019} into the \emph{Deep InfoMax} principle (DIM). Instead of focussing on a layer-wise maximisation of mutual information, the key proposal is to both maximise the mutual information between data and clusters as we previously described, but also enforce high mutual information between subsets of visual features of an image and the clusters. Denoting $g_{\theta_1}$ the local feature learning function and $\psi_{\theta_2}$ the representation function, the DIM can be written as follows:
\begin{equation}
    \theta_1^\star, \theta_2^\star \in \argmax_{\theta_1,\theta_2} \mi{\x; \psi_{\theta_2} \circ g_{\theta_1} (\x)} + \lambda \sum_{i=1}^M \mi{g_{\theta_1}(\x)^{(i)}; \psi_{\theta_2} \circ g_{\theta_1} (\x)},
\end{equation}

\noindent where $g_{\theta_1}^{(i)}$ is the $i$-th subset of $M$ subsets of features, \eg some pixels in an image. Notice that we omitted the prior matching constraint of \citet{hjelm_learning_2019} for clarity. In this original work, the NCE estimator was used for mutual information~\citep{van_den_oord_representation_2018}, see Eq.~(\ref{eq:info_nce}).

The DIM framework has been extended, for example, to focus beyond local subsets of data features and instead consider augmentations of the data~\citep{bachman_learning_2019}. It was also incorporated into GANs~\citep{lee_infomax-gan_2021} to avoid mode collapse. Nonetheless, we find that the modern meaning of InfoMax differs from mutual information for clustering as seen in the previous sections.

\subsection{Dissonance between MI and performances}
\label{ssec:thrive_limitations}

Evaluating the mutual information between two continuous random variables is challenging due to the intractability of the underlying integrals. Therefore, lower bounds were derived to alleviate the maximisation of mutual information. However, these bounds can come with either high variance or high bias. That is why \citet{poole_variational_2019} proposed an interpolated lower bound to offer a trade-off between variance and bias: $\mathcal{I}_\alpha$. Nonetheless, it was noticed that MI is hardly predictive of downstream tasks~\citep{tschannen_mutual_2020} with the output continuous variable $\z$. In other words, a high value of mutual information does not clarify whether the learnt continuous representations are insightful and can leverage a second-step task such as clustering or classification. This joins the criticism on mutual information directly maximised for clustering that we discussed in Section~\ref{ssec:challenge_mutual_information}. Additionally, \citet{mcallester_formal_2020} proved that any lower bound on mutual information with $n\geq 50$ samples per batch cannot be greater than $\mathcal{O}(\log n)$. To conclude on mutual information, especially in contrastive models, since its value is constant and the model is often an ernegy-based variational approximation of $\p(\x\mid \z)$, perhaps a revision of what the objective actually is should be done. In this sense, \citet{mcallester_formal_2020} presented an estimation of mutual information via lower bound on entropies, even though their proposal might neither be an upper bound nor a lower bound on MI.

These observations stands in contrast to the most recent articles on deep clustering that achieve good performance in supervised datasets, \eg CIFAR10 and ImageNet~\citep{lee_unsupervised_2022, park_improving_2021, dang_nearest_2021}. Interestingly, the datasets used for benchmarking often focus on images and rarely other types of data such as tabular data~\citep{min_survey_2018}. Overall, it is plausible that the success of these methods may be due to the good design of the discriminative model's architecture which encompasses the underlying assumptions and its regularisations, rather than mutual information itself. \citet{ren_deep_2024} noted that:

\textit{``Due to the complexity brought by massive data, most of the existing deep clustering models are designed for specific data sets. Complex data from different sources and forms bring more uncertainties and challenges to clustering.''}

Consequently, we empirically observe that the number of clusters to find is an often overlooked question when the evaluation protocol lies on datasets in which the number of \emph{classes}, not clusters, is known. Therefore, an interesting clustering algorithm should be able to find a relevant number of clusters, \ie perform model selection. However, model selection for parametric deep clustering models is expensive~\citep{ronen_deepdpm_2022}.

\section{Model selection in discriminative clustering}
\label{sec:model_selection}

We have seen so far that discriminative clustering initially started with a decoupling between the model, \ie the type of decision boundary that we allow between clusters, and the objective function to optimise the model's parameters: mutual information. With the thrive of self-supervised learning, and especially contrastive learning, discriminative clustering took a turn where the model is no longer a decision boundary in the data space, but a critic function to identify samples from a common neighbourhood. Therefore, maximisation of mutual information in a contrastive fashion enforces a different interpretation on the nature of clusters: model and objective become entangled. The clustering hypotheses now concern the nature of neighbourhoods: we search how many augmentations away a sample is still the same sample. Finally, we further saw that mutual information was an objective that required regularisations to become a relevant discriminative objective, both for categorical and continuous predictions.

Once the hypotheses on the nature of the clusters are clearly defined, the second question to address is the number of clusters. To that end, we need \emph{internal} metrics, \ie scores that work on at most two inputs: the model's parameters and its predictions on the dataset. In contrast, unsupervised learning external metrics like the adjusted rand index (ARI~\citealp{hubert_comparing_1985}) require external labels that tell us how correct our clustering is. While it makes sense in a synthetic case where we have knowledge of what the model should discover and where we control the definition of the clusters, this is inapplicable to \emph{true} unsupervised cases. In an exploratory scenario, we do not have access to any ground truth. That is why model selection must not rely on external metrics.

\subsection{Existing internal metrics}

In discriminative clustering, there exist internal metrics that are based on the model's predictions over the dataset: the K-means score, the silhouette score~\citep{rousseeuw_silhouettes_1987}, the Davies-Bouldin index~\citep{davies_cluster_1979}, or the Dunn index~\citep{dunn_well-separated_1974} for example. As we covered the K-means score in Section~\ref{sec:other}, here we detail the silhouette score. For each sample $\x$ belonging to the cluster $k$, we start by computing its average intra-cluster distance:
\begin{equation}
\text{Intra}(\x) = \inv{\card{\cluster_k}-1} \sum_{\x^\prime \in \cluster_k \setminus \x} c(\x, \x^\prime),
\end{equation}

\noindent and its minimal average outer-cluster distance:
\begin{equation}
    \text{Outer}(\x) = \min_{k^\prime \neq k} \inv{\card{\cluster_{k^\prime}}} \sum_{\x^\prime \in \cluster_{k^\prime}} c(\x, \x^\prime),
\end{equation}

\noindent where $c$ is the distance or cost of moving $\x$ to $\x^\prime$. Note that both terms are sometimes referred to as average similarities. The silhouette score of the sample $\x$ is then the ratio between the difference of both terms divided by the greatest:
\begin{equation}
\text{Silhouette}(\x) = \frac{\text{Outer}(\x) - \text{Intra}(\x)}{\max \{ \text{Intra}(\x), \text{Outer}(\x)\}}.
\end{equation}

The global silhouette score used to describe a model is the average between the individual silhouette values of each sample. In the same spirit, other scores like the Dunn index or the Davies-Bouldin index employ a notion of distance $c$.

To select a number of clusters, the common practice for the K-means score is to plot its value for all models trained on an increasing number of clusters. Finding the spot where the curve bends the most, called elbow, is expected to highlight a relevant number of clusters. However, such an approach can be criticised or discouraged due to the absence of proper definition of an elbow~\citep{schubert_stop_2023}, leading to unclear values of the number of clusters to retain. Instead, statistical methods that compare the score decrease on the dataset against bootstrap samples can be preferred, such as the gap statistic~\citep{tibshirani_estimating_2001}. When the score decreases in a greater manner than its expectation over bootstrap samples, perhaps the model did find a suitable decision boundary. For other internal metrics, such as the Silhouette score, finding the highest value is common practice to choose the number of clusters.

This is just a brief overview of the existing metrics. \citet{vendramin_relative_2010} compared for instance 40 different internal criteria for assessing clustering quality, a majority of which being available in Python for instance with the permetrics library\citep{thieu_permetrics_2024}. 

\subsection{The necessity of adapting the internal metric to the model}

There is however a critical dissonance in using such metrics in a raw manner: they do not respect the nature of the clustering hypotheses. To illustrate that point, let us take a kernelised model:
\begin{equation}
\p(y = k\mid \x) \propto \exp \left(\sum_{i=1}^n a_{ik} \kappa (\x_i, \x)\right),
\end{equation}

\noindent where $\kappa$ is a kernel tied to an RKHS $\rkhs$ and the parameters $a_{ik}$ weight the kernel terms between the sample $\x$ and the i-th dataset sample $\x_i$ for the $k$-th cluster.  Such a model can be trained by maximising mutual information, which is known as kernel RIM~\citep{krause_discriminative_2010}. This model seeks a linear boundary in a high-dimensional space. If the kernel is Gaussian, that space has infinite dimension. If the kernel is linear, then that space has the same dimension has $\dataspace$. This consequently means that the model seeks to have a maximal distance between sample in \emph{this} high-dimensional space, rather than the initial data space. Consequently, evaluating the clustering performances of this model using for instance a silhouette score or a kernel K-means that were not adequately changed to take into account the kernelised nature of the data will mislead the conclusions.

This reasoning can be extended to most cases where we seek decision boundaries that are nonlinear between the samples: internal scores based on distances must be changed. Moreover, when the nature of hypotheses diverges from decision boundaries to invariances, perhaps distance-based internal scores should be replaced by more relevant internal scores.

This limitation stands in contrast to generative scores, \eg ICL or AIC. Indeed, generative scores for model selection can incorporate the likelihood value, therefore tacitly encompassing the generative hypotheses of the model within their definition. Discriminative scores rather come with their own set of hypotheses on the nature of clusters in a post hoc fashion, rather than being based on the definition of the clustering model.

A potential solution to address this duality would be to incorporate the model selection mechanism within the model. For example, \citet{ronen_deepdpm_2022} proposed a mechanism of cluster merging and splitting within their deep network architecture inspired by the Dirichlet process Gaussian mixture models (DPGMM, \citealp{antoniak_mixtures_1974}). Using a Hastings ratio to accept split or merge proposals, they progressively duplicate or create new neurons in the clustering layer. The architecture is composed of 3 main parts: a feature extractor, which could be trained using contrastive learning techniques, a main clustering layer, and sub-clustering multi-layered perceptrons that break down each cluster in two parts. Thus, without having any prior knowledge on the number of clusters ahead of training, the model can be tuned during optimisation to perform selection.

It is important to note that in recent surveys~\citet{ren_deep_2024, min_survey_2018, zhou_comprehensive_2022} for deep clustering, model selection is often overlooked. Some of them mention the question of selecting the right number of clusters but not deeply explore this challenging issue. \citet{wei_overview_2024} give some examples of deep learning models that successfully integrate mechanisms for selecting a number of cluster. These surveys encompass both generative and discriminative clustering models.

\section{A complete example}
\label{sec:example}
Let us give a complete example on how to use some discriminative clustering models. We detail this entire example with matching code snippets for reproducibility. The code of this section can also be found in the following companion notebook: \url{https://github.com/gemini-clustering/A-tutorial-on-discriminative-clustering}.

We solve this example with distinct parts. In the first part, we use models that define $p(y\mid \x)$, using the packages NumPy v1.26.4, Scikit-Learn v1.5.2 and GemClus v1.0.0. This covers the sections~\ref{ssec:linear_boundary_models}, \ref{ssec:projection_models} and \ref{ssec:small_neural_networks}. In the second part, Section~\ref{ssec:contrastive_clustering}, we identify the clusters using a contrastive model with PyTorch v2.5.0. No CUDA acceleration is needed.

\subsection{GemClus: a package for discriminative clustering}

Among the packages cited above, GemClus is a specialised package for discriminative clustering models. It was initially built around models trained using GEMINI~\citep{ohl_generalised_2022, ohl_sparse_2024}, see Eq.~(\ref{eq:gemini}). Today, it is extended and welcomes any discriminative clustering methods, \eg RIM~\cite{krause_discriminative_2010}, logistic regression with feature selection using MI~\cite{kong_discriminative_2015}. An up-to-date list of implemented methods is available at: \url{https://gemini-clustering.github.io/main/user_guide.html#a-summary-of-what-is-implemented}. The package aims at keeping a decoupling between a choice of architecture, \ie decision boundary, and the training objective, as discussed in Section~\ref{ssec:thrive_beginning}.

\subsection{The dataset}

\begin{lstlisting}[language=PythonPlus, style=colorEX,caption={Sampling of the circle dataset.}, label={list:concentric_circles}, float, floatplacement=t]
from sklearn import datasets
X, y = datasets.make_circles(n_samples=200, noise=0.05, factor=0.1, random_state=0)
X = (X-X.mean(axis=0))/X.std(axis=0)
\end{lstlisting}

For this example, we will focus on a simple dataset consisting of two concentric circles, also known as two rings, using Listing~\ref{list:concentric_circles}.

\subsection{Linear boundary models}
\label{ssec:linear_boundary_models}

In a generative modelling perspective, we would assume that data is distributed according to two concentric circles. The challenge here would lie in building such nonlinear distributions. In a discriminative clustering perspective, assumptions are made on the shape of the decision boundary. We can start with linear decision boundaries. It is clear that they will not be efficient for these datasets, as it is impossible to separate the two circles with a straight line. Consequently, models such as KMeans (see Section~\ref{ssec:other_kmeans}) or RIM~\cite{krause_discriminative_2010} (see Section~\ref{ssec:thrive_beginning}) are expected to not perform well. We give a snippet of code in Listing~\ref{list:linear_boundary_models} where KMeans clustering gets an ARI of 0.178 and RIM an ARI of -0.003.

For KMeans, the model is written as follows:
\begin{equation}
    \p(y=k\mid \x) = \ind[k = \argmin_{k^\prime} \norm{\x - \vec{\mu}_{k^\prime}}_2^2],
\end{equation}

\noindent where $\theta=\{\vec{\mu}_k\}, \mu_k \in \mathbb{R}^2$ are the centroid parameters. For RIM, it is a logistic regression model:
\begin{equation}
    \p(y\mid \x) = \text{SoftMax}(\vec{W}^\top\x + \vec{b}),
\end{equation}

\noindent where $\theta=\{\vec{W}, \vec{b}\}$, and $\vec{W} \in \real^{2\times 2}$ and $\vec{b} \in \real^2$ are the parameters.

\begin{lstlisting}[language=PythonPlus, style=colorEX,caption={Clustering of circle dataset by models with linear decision boundaries.}, label={list:linear_boundary_models}, float, floatplacement=t]
from sklearn import cluster, metrics
import gemclus
# Create a KMeans model
kmeans_model = cluster.KMeans(n_clusters=2, random_state=0)
# Fit to the data. Note the absence `y`in the fit function.
y_kmeans = kmeans_model.fit_predict(X)
# Evaluate ARI with known targets.
ari_kmeans = metrics.adjusted_rand_score(y, y_kmeans)
print(f"ARI of KMeans predictions: {ari_kmeans:.3f}") # 0.174
# Create a linear RIM clustering model
linear_rim_model = gemclus.linear.RIM(n_clusters=2, random_state=0)
y_linear_rim = linear_rim_model.fit_predict(X)
ari_linear = metrics.adjusted_rand_score(y, y_linear_rim)
print(f"ARI of Linear RIM predictions: {ari_linear:.3f}") # -0.003
\end{lstlisting}

\subsection{Projection models}
\label{ssec:projection_models}

To enhance the performances for this dataset, we have a visual cue that a straight line is not sufficient to separate both circles. A first approach we can try in this context is to project the dataset into a space where it is easier to separate. We can do this with both Spectral clustering, see Section~\ref{ssec:other_spectral}, or kernel RIM, see Section~\ref{ssec:thrive_beginning}. The spectral clustering is trained using the default parameters of scikit-learn. For kernel RIM, we use a radial-basis function kernel, which is known to produce a satisfying space for separating these circles, and no $\ell_2$ penalty. 

We can summarise the spectral model with:
\begin{equation}
    \p(y=k\mid \x) = \ind[k = \argmin_{k^\prime} \norm{\gamma(\x) - \vec{\mu}_{k^\prime}}_2^2],
\end{equation}
\noindent where $\gamma$ is the spectral embedding of $\x$ after finding the $K$ largest eigenvalues of the proximity matrix of the dataset. This novel space is of dimension 2 because we seek two clusters. The parameters are therefore both $\vec{\mu}_k$ due to the final KMeans clustering algorithm, and $\gamma$, which hides assumptions on the relevant notion of proximity.

For kernel RIM, the model is written as follows:
\begin{equation}
    \p(y \mid \x) = \text{Softmax}(\vec{W}^\top\vec{\kappa}(\x) + \vec{b})
\end{equation}

\noindent where $\kappa(\x) = [\kappa(\x_i, \x)] \in \real^n$ is the vector that describes the kernel value between the sample $\x$ and all dataset samples. Consequently, the parameters are $\vec{W} \in \real^{n\times 2}$, $\vec{b} \in \real^2$, and the choice of a kernel function $\kappa$.

\begin{lstlisting}[language=PythonPlus, style=colorEX,caption={Clustering of circle dataset by projection/kernel based models.}, label={list:kernel_models}, float, floatplacement=t]
# Create a spectral clustering model
spectral_model = cluster.SpectralClustering(n_clusters=2)
y_spectral = spectral_model.fit_predict(X)
ari_spectral = metrics.adjusted_rand_score(y, y_spectral)
print(f"ARI of Spectral predictions: {ari_spectral:.3f}") # 1.000
# Create a kernel RIM clustering model without l2 penalty
kernel_rim_model = gemclus.linear.KernelRIM(n_clusters=2, base_kernel="rbf", reg=0, random_state=0)
y_kernel_rim = kernel_rim_model.fit_predict(X)
ari_kernel = metrics.adjusted_rand_score(y, y_kernel_rim)
print(f"ARI of Kernel RIM predictions: {ari_kernel:.3f}") # 1.000
\end{lstlisting}

We can then seek to enhance the properties of the algorithms. For instance, spectral clustering cannot generalise. It cannot assign a clustering probability to any sample outside of the training dataset. This means that we cannot observe the decision boundary of this algorithm. Generalising can be a desirable property, for example when we want to evaluate the conditional entropy around the decision boundary or find adversarial examples against the model. In the case of kernel RIM, although generalisation is possible, it requires to compute the kernel with all samples from the training set. This implies that the generalisation will scale linearly with the number of samples and may be expensive for large training datasets.

\subsection{Small neural networks}
\label{ssec:small_neural_networks}

Neural networks can simultaneously generalise and derive a projection of the dataset. 
In this case, a small neural network will suffice: we just need a single hidden layer with a couple nodes to place hyperplanes in a circular manner. Then, the final layer will draw the clustering hyperplane in the hidden representation. In the example of Listing~\ref{list:neural_networks}, we train 2 neural networks with 20 hidden nodes. The first one is trained using MI. However, due to a lack of regularisation on a flexible architecture, this model does not perform well and gets an ARI of 0.278. This follows the critics on MI we discussed in Section~\ref{ssec:challenge_mutual_information}. In contrast, the second model is trained with GEMINI, a distance-based generalisation of MI where the involvement of a distance in the objective acts as a regulariser. As we did with kernel RIM, we use an RBF kernel in the objective of this model. This model achieves perfect ARI.

For both models, the underlying architecture is:
\begin{equation}
    \psi_\theta(\x) = \vec{W}_2^\top \text{ReLU}(\vec{W}_1^\top \x + \vec{b}_1) + \vec{b}_2.
\end{equation}

\begin{lstlisting}[language=PythonPlus, style=colorEX,caption={Clustering of circle dataset by neural network-based models.}, label={list:neural_networks}, float, floatplacement=t]
# Create a neural network with MI objective
mlp_mi_model = gemclus.mlp.MLPModel(n_clusters=2, gemini="mi", n_hidden_dim=20, random_state=0)
y_mlp_mi = mlp_mi_model.fit_predict(X)
ari_mlp_mi = metrics.adjusted_rand_score(y, y_mlp_mi)
print(f"ARI of MLP MI predictions: {ari_mlp_mi:.3f}") # 0.278
# Create a neural network with GEMINI objective
mlp_gemini_model = gemclus.mlp.MLPMMD(n_clusters=2, n_hidden_dim=20, kernel="rbf", random_state=0)
y_mlp_gemini = mlp_gemini_model.fit_predict(X)
ari_mlp_gemini = metrics.adjusted_rand_score(y, y_mlp_gemini)
print(f"ARI of MLP MMD-GEMINI predictions: {ari_mlp_gemini:.3f}") #1
\end{lstlisting}

\begin{figure}[t]
    \centering
    % \subfloat[KMeans]{
    %     \includegraphics[width=0.3\linewidth]{imgs/circles_kmeans.pdf}
    % }\hfill
    % \subfloat[Spectral]{
    %     \includegraphics[width=0.3\linewidth]{imgs/circles_spectral.pdf}
    % }\hfill
    % \subfloat[MLP+MI]{
    %     \includegraphics[width=0.3\linewidth]{imgs/circles_mlp_mi.pdf}
    % }\\
    % \subfloat[RIM]{
    %     \includegraphics[width=0.3\linewidth]{imgs/circles_linear_rim.pdf}
    % }\hfill
    % \subfloat[Kernel RIM]{
    %     \includegraphics[width=0.3\linewidth]{imgs/circles_kernel_rim.pdf}
    % }\hfill
    % \subfloat[MLP+GEMINI]{
    %     \includegraphics[width=0.3\linewidth]{imgs/circles_mlp_gemini.pdf}
    % }
    \subfloat[KMeans]{
        \includegraphics[width=0.25\linewidth]{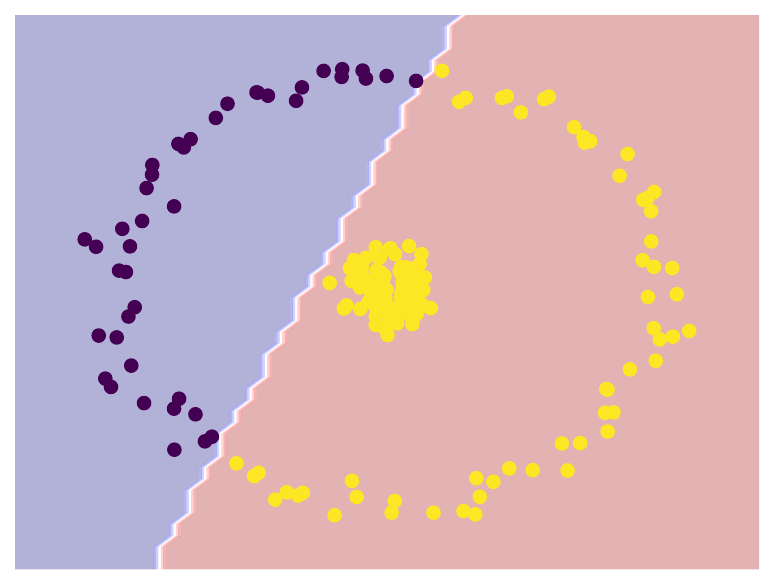}
    }\hspace{1em}
    \subfloat[Spectral]{
        \includegraphics[width=0.25\linewidth]{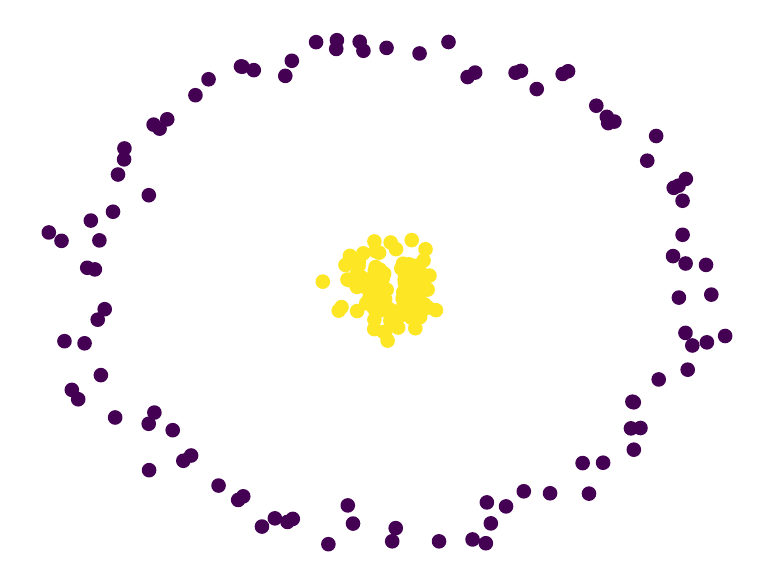}
    }\hspace{1em}
    \subfloat[MLP+MI]{
        \includegraphics[width=0.25\linewidth]{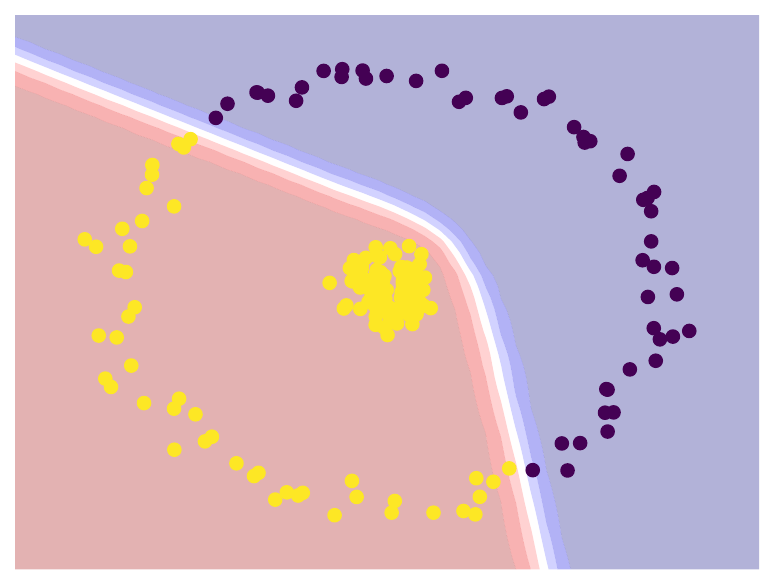}
    }\\
    \subfloat[RIM]{
        \includegraphics[width=0.25\linewidth]{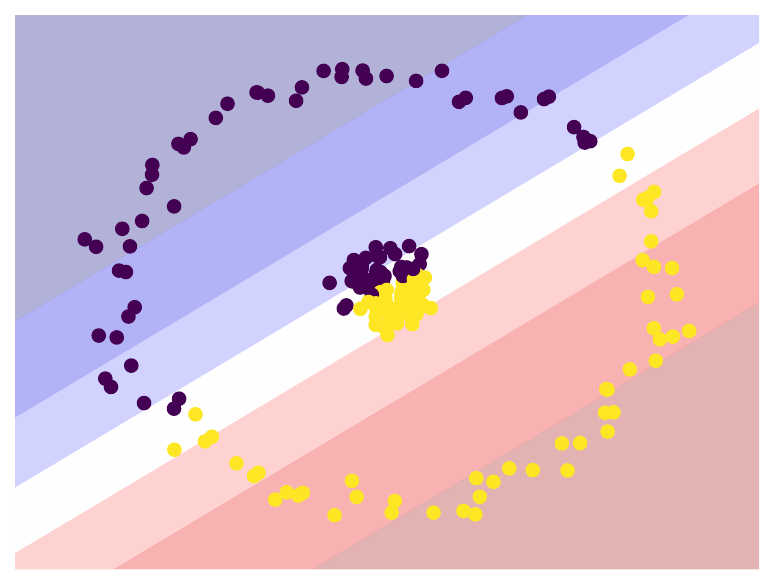}
    }\hspace{1em}
    \subfloat[Kernel RIM]{
        \includegraphics[width=0.25\linewidth]{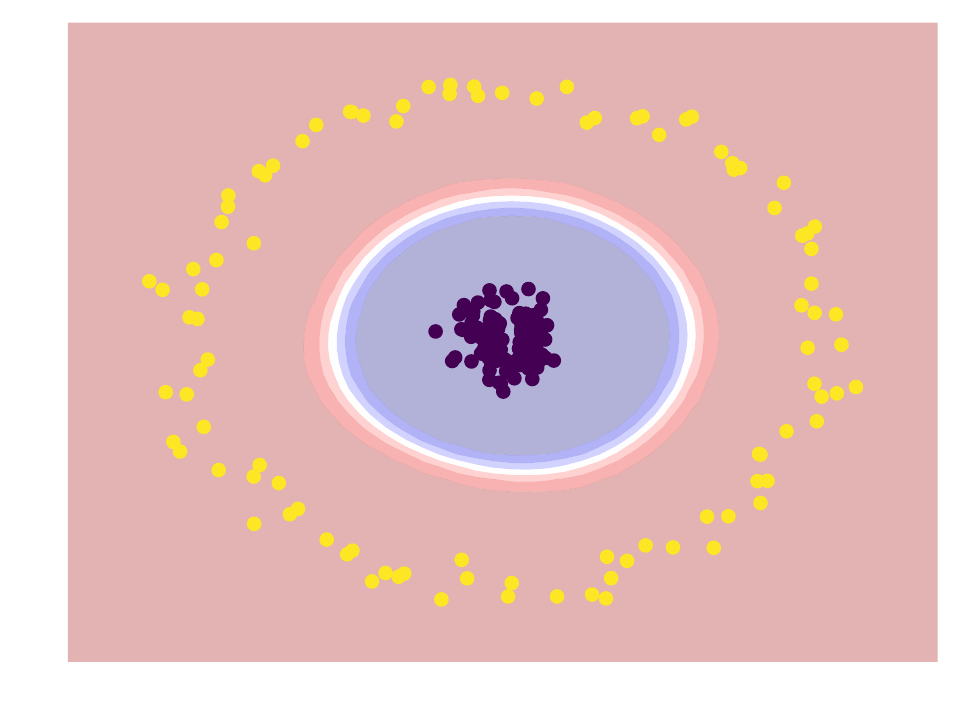}
    }\hspace{1em}
    \subfloat[MLP+GEMINI]{
        \includegraphics[width=0.25\linewidth]{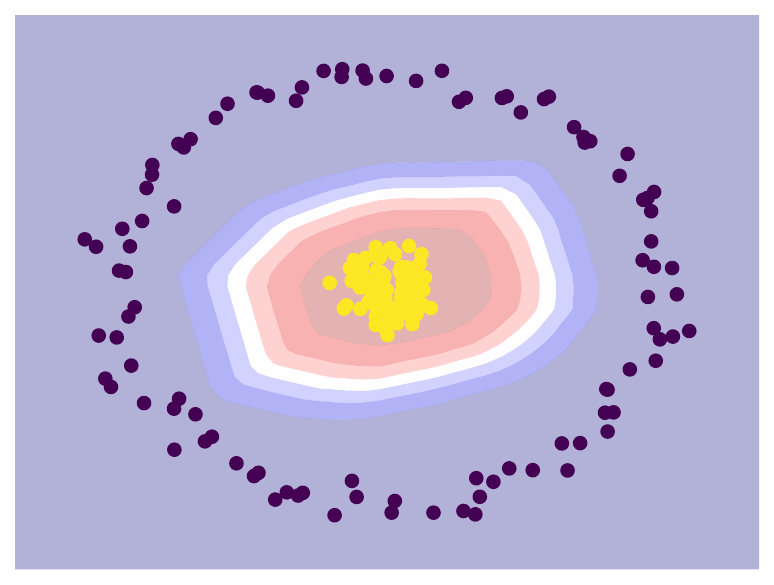}
    }
    \caption{Final clustering and decision boundaries of the example models on a circle dataset. Blue stands for low probability of cluster 2, red for high.}
    \label{fig:final_circle_clustering}
\end{figure}

We can observe in Figure~\ref{fig:final_circle_clustering} the final clusterings obtained by each model in this example. For the models that can generalise, \ie all but spectral clustering and kernel RIM, we plot the decision boundary by showing the probability of cluster 2.

\subsection{Identifying clusters with contrastive learning}
\label{ssec:contrastive_clustering}

For completeness of the example, we now expand the methods by showing how the clusters can be found using a contrastive learning method.

If we want to tackle the concentric circles using contrastive learning, we need to think in terms of invariances, as discussed in Section~\ref{ssec:thrive_contrastive}. This invariance will guide the choice of augmentations we can employ. For this dataset, it is clear that there is an invariance to rotation: no matter the angle of rotation around the origin, the clusters we aim for remain the same. Consequently, a rotation-based augmentation with a random angle may be a good strategy for using a contrastive loss. In contrast, adding random Gaussian noise may not be a good strategy as it will not necessarily suffice to make a good distinction between inner and outer circles. Let us compare both.

We take the exact same neural network architecture as before: a single hidden layer with 20 nodes. The only difference we may have is that there will not be a softmax activation at the end of this neural network. Indeed, the goal of this network is to discriminate between the respective augmented samples and the remaining samples, recalling Eq.~(\ref{eq:ntxent}). In other words, this network does not define a clustering distribution. In this example, we use the cosine similarity between two vectors. This implies that we need to normalise the output representations:
\begin{equation}
    E_\text{contrastive}(\x_i, \text{Aug}(\x_j)) = \left\langle \frac{\psi_\theta(\x_i)}{\norm{\psi_\theta(\x_i)}}, \frac{\psi_\theta(\text{Aug}(\x_j))}{\norm{\psi_\theta(\text{Aug}(\x_j))}}\right\rangle.
\end{equation}

The model is:
\begin{equation}
    q_\theta(\x \mid \text{Aug}(\x)) \propto \pdata(\x) e^{E_\text{contrastive}(\x, \text{Aug}(\x))},
\end{equation}
and we optimise its parameters $\theta$ such that $q_\theta$ approximates $p(\x\mid \text{Aug}(\x))$.

To extract a clustering out of this model, we choose to have 2 output dimensions, 1 per cluster, and will take the argmax as the final cluster, following the examples from Section~\ref{sssec:contrastive_learning}. We give with Listing~\ref{list:contrastive_example} an example of training such a model using a random rotation per batch in $[0, 2\pi]$ during 5000 epochs and a learning rate of $10^{-4}$ in an Adam optimiser.

\begin{lstlisting}[language=PythonPlus, style=colorEX,caption={Clustering of circle dataset using invariances in contrastive learning.}, label={list:contrastive_example}, float, floatplacement=t]
import torch
torch.manual_seed(0)
model = torch.nn.Sequential(
    torch.nn.Linear(X.shape[1], 20),
    torch.nn.ReLU(),
    torch.nn.Linear(20, 2) # Observe the absence of softmax
)
optimiser = torch.optim.Adam(model.parameters(), lr=1e-4)
X = torch.Tensor(X)
for i in tqdm(range(5000)):
    # AUGMENTATION
    theta = 2*torch.pi*torch.rand(1) # [0, 2*pi]
    rotation_matrix = torch.Tensor(
        [[torch.cos(theta), -torch.sin(theta)],
        [torch.sin(theta), torch.cos(theta)]])
    augmented_samples = X@rotation_matrix.T
    # PREDICTION
    original_prediction = model(X)
    with torch.no_grad():
        # For simplicity, we do not backpropagate on augmented samples
        augmented_prediction = model(augmented_samples)
    # LOSS
    ## Softmax of cosine similarity energy
    original_normalised = original_prediction /
        torch.norm(original_prediction, dim=1, keepdim=True)
    augmented_normalised = augmented_prediction /
        torch.norm(augmented_prediction, dim=1, keepdim=True)
    similarities = original_normalised @ augmented_normalised.T
    loss = torch.diag(torch.softmax(similarities, dim=0))
    loss = -loss.sum() ## Negate because we seek to maximise
    # GRADIENT DESCENT
    optimiser.zero_grad()
    loss.backward()
    optimiser.step()
\end{lstlisting}

Exactly as in the previous case, using a neural network allows us to generalise. However, we must keep in mind that the predictions we use here to draw a decision boundary correspond to the highest value of a 2-dimensional vector intended to discriminate between augmented and original sample pairs. The plotted values do not therefore represent a clustering conditional probability $\p(y\mid\x)$.

\begin{figure}[t]
    \centering
    \subfloat[$\x + \mathcal{N}(0, 1)$]{
        \includegraphics[width=0.25\linewidth]{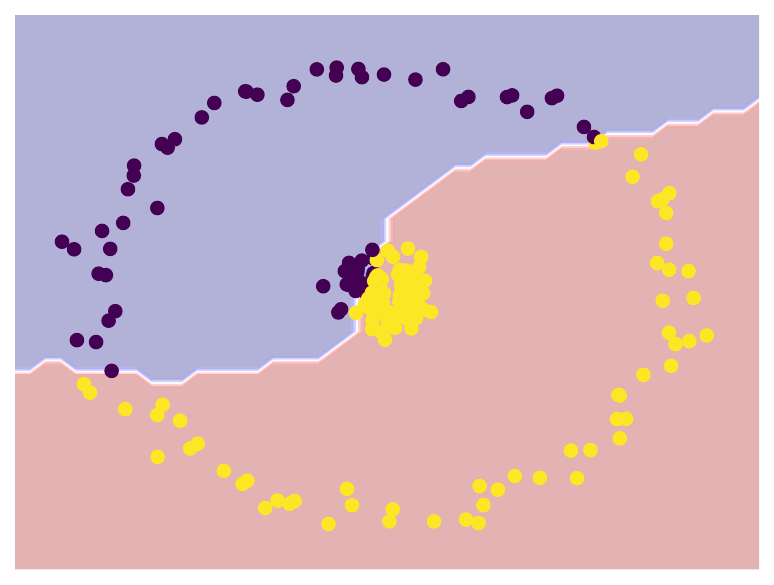}
    }\hspace{1em}
    \subfloat[$\x + \mathcal{N}(0, 0.01^2)$]{
        \includegraphics[width=0.25\linewidth]{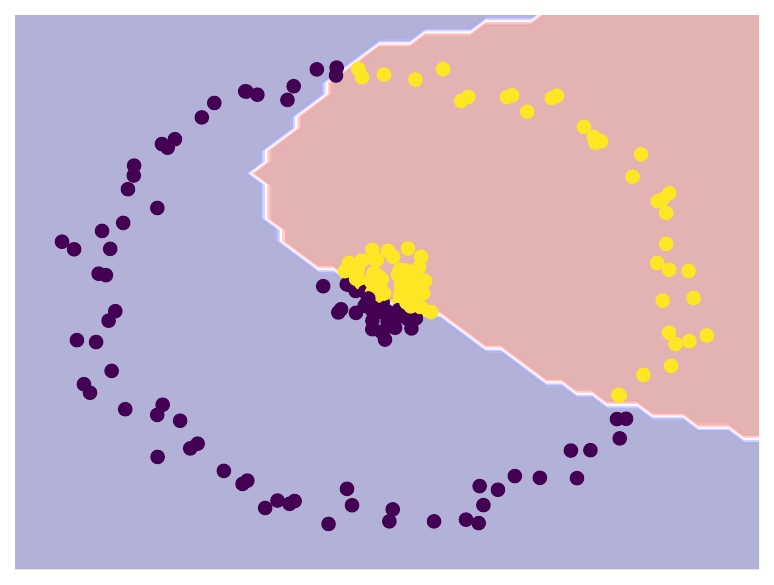}
    }\hspace{1em}
    \subfloat[$\text{Rotation}_{[0, 2\pi]}(\x)$]{
        \includegraphics[width=0.25\linewidth]{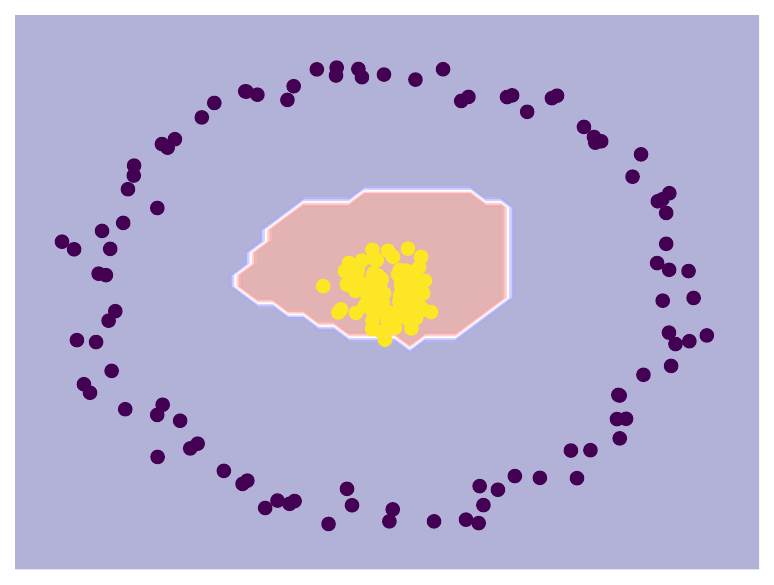}
    }\\
    \caption{Decision boundary using the maximal dimension of the contrastive critic function depending on the choice of augmentation on the data. We either add Gaussian noise or randomly rotate the dataset around the origin.}
    \label{fig:circle_invariance_boundaries}
\end{figure}

\begin{figure}[t]
    \centering
    \subfloat[$\x + \mathcal{N}(0, 1)$]{
        \includegraphics[width=0.25\linewidth]{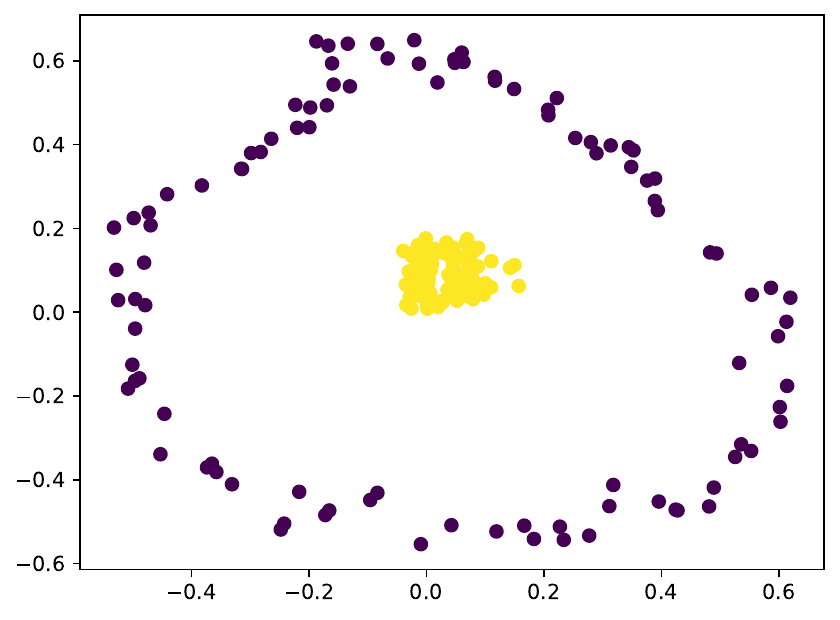}
    }\hspace{1em}
    \subfloat[$\x + \mathcal{N}(0, 0.05^2)$]{
        \includegraphics[width=0.25\linewidth]{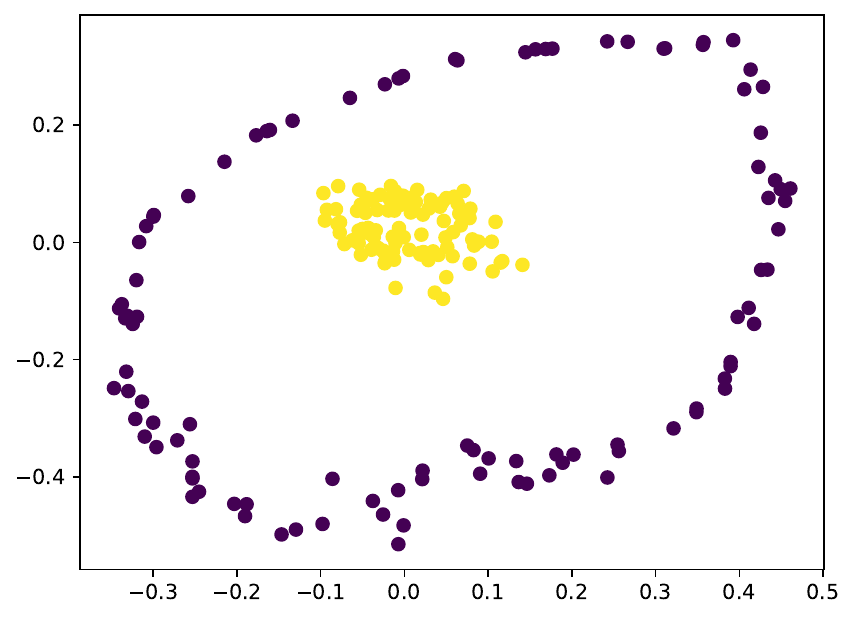}
    }\hspace{1em}
    \subfloat[$\text{Rotation}_{[0, 2\pi]}(\x)$]{
        \includegraphics[width=0.25\linewidth]{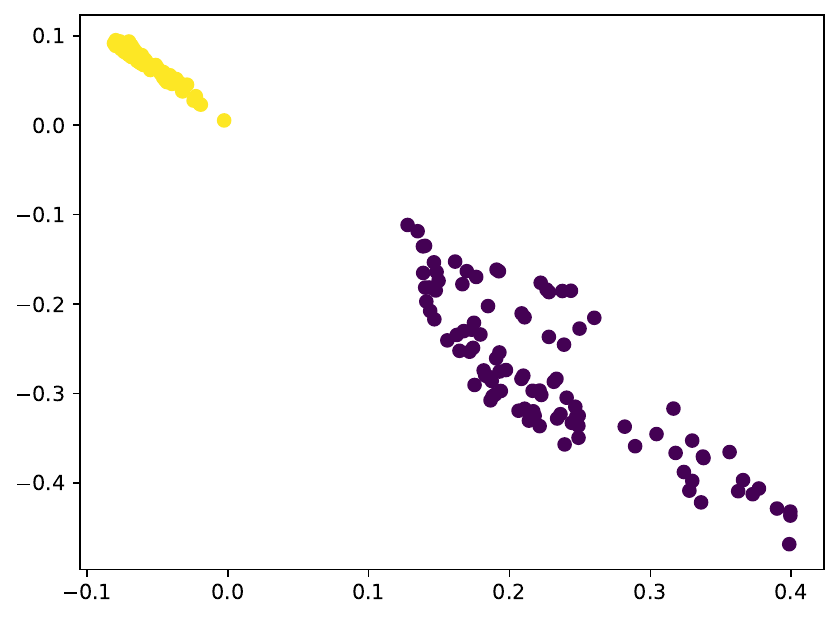}
    }\\
    \caption{Unnormalised 2d intermediate representation of the dataset produced by the discriminating neural network. Decision boundary is made depending on the location above or below the identity axis. Samples are coloured according to inner/outer circles.}
    \label{fig:circle_intermediate_contrastive}
\end{figure}

Using the same parameters and initial weights, we can compare the decision boundary obtained by this model depending on the augmentation strategy. We show the different boundaries in Figure~\ref{fig:circle_invariance_boundaries}, and the matching 2d representation returned by the model in Figure~\ref{fig:circle_intermediate_contrastive}. When adding unit Gaussian noise, the scale of the noise overlaps all the dataset. Consequently, all samples are considered close to each other and a suitable choice of decision boundary is to cut in the middle. Reducing the noise scale does not solve the issue.

Rotation invariance may not be sufficient for more challenging datasets. If we generalise to multiple circles, rotation would not guarantee a one-to-one relationship between clusters and circles: a model could put two circles in the same cluster, and would remain invariant to rotation. So ideally, we should seek a combination of a small-scale Gaussian noise and rotation to have both invariance to small radius change and angle variations.

One final criticism we can elaborate on this example is that we \emph{knew} the form of the clusters we were looking for. This means that we rigged the choice of augmentations, distance or model to achieve the desired clusters. In a data analysis context, we would not necessarily have such a "prior" on the choice of models for clustering, and expert knowledge would be expected to build reasonable hypotheses on invariances or decision boundary shapes.

\section{Take-aways}

We presented here an overview of the historical evolution of discriminative clustering and how mutual information was a cornerstone of its development. We started by detailing how the discriminative clustering assumptions are different from generative models that assume probability distributions on the data. We showed that the initial hypotheses focused more on using a specific type of decision boundary with shallow models. Then, we saw how that notion of decision boundary was dropped by the usage of nonparametric models in different fashions, especially with neural networks growing deeper and deeper. Finally, we observed how the nature of the discriminative hypotheses changed to defining neighbourhoods of samples through invariances and augmentations to construct clusters. The ensuing objectives, lower bounds on mutual information, then became entangled with the model in the clustering hypotheses. Consequently, models trained today rather correspond to critic functions that aim at identifying similar samples, rather than a straightforward distribution $\p(y\mid \x)$. We notably highlighted that the objective is the lower bound on a constant value of mutual information, and that the true underlying model is an energy-based model aiming at recovering true samples $\x$ given an augmentation $\z$: $q_\theta(\x\mid \z)$.

Concentrating on mutual information, we covered some of its limitations that are often depicted by related works. We observed that the rise of mutual information in discriminative clustering was therefore accompanied by increases in the attempts to regularise its limitations: weight decay, adversarial penalty, contrastive invariances. It is likely that the recent successes of deep clustering methods using mutual information should be granted to the clever design of such regularisations, rather than the objective itself. Moreover, the identification of the actual energy-based model in contrastive learning questions whether mutual information is actually the objective we need to learn deep clustering method, or whether novel objectives should be considered for optimising their parameters.

Finally, we observed that discriminative clustering, especially regarding deep methods, calls for an adaptation of model selection strategies. The internal metrics should typically be in line with the clustering hypotheses that are used to construct clustering instead of being post hoc tools in an evaluation protocol. The crucial questions of clustering revolve both on clustering hypotheses and their consequential impact on what could be deemed a good number of clusters. In this sense, incorporating model selection mechanisms within the architecture of the training procedure could be a promising direction.

\bibliographystyle{plainnat}
\bibliography{bib}

\end{document}